\documentclass[a4paper,fleqn]{cas-dc}
\usepackage{bm}
\usepackage[numbers]{natbib}
\usepackage{makecell}
\usepackage{hyperref}
\def\tsc#1{\csdef{#1}{\textsc{\lowercase{#1}}\xspace}}
\tsc{WGM}
\tsc{QE}
\tsc{EP}
\tsc{PMS}
\tsc{BEC}
\tsc{DE}

\begin{document}
\let\WriteBookmarks\relax
\def\floatpagepagefraction{1}
\def\textpagefraction{.001}

\shorttitle{Surgical-VQLA++: Adversarial Contrastive Learning for Calibrated Robust VQLA in Robotic Surgery}

\shortauthors{Long Bai et~al.}

\title [mode = title]{Surgical-VQLA++: Adversarial Contrastive Learning for Calibrated Robust Visual Question-Localized Answering in Robotic Surgery}                     

\author[1,2]{Long Bai}[orcid=0000-0002-9762-6821]
\fnmark[1]

\ead{b.long@link.cuhk.edu.hk}
\credit{Conceptualization, Methodology, Validation, Investigation, Original draft preparation}
\affiliation[1]{organization={Department of Electronic Engineering, The Chinese University of Hong Kong},
    city={Hong Kong},
    country={China}}
\affiliation[2]{organization={Shenzhen Research Institute, The Chinese University of Hong Kong},
    city={Shenzhen},
    country={China}}

\author[1,2]{Guankun Wang}[orcid=0000-0003-2440-4950]
\fnmark[1]
\ead{gkwang@link.cuhk.edu.hk}

\credit{Methodology, Validation, Investigation, Original draft preparation}

\author[3]{Mobarakol Islam}[orcid=0000-0002-7162-2822]
\fnmark[1]
\ead{mobarakol.islam@ucl.ac.uk}

\credit{Conceptualization, Methodology, Investigation, Original draft preparation}

\affiliation[3]{organization={Wellcome/EPSRC Centre for Interventional and Surgical Sciences (WEISS), University College London},
    city={London},
    country={UK}}

\author[1, 4]{Lalithkumar Seenivasan}[orcid=0000-0003-4312-7151]

\ead{lalithkumar_s@u.nus.edu}

\credit{Investigation, Methodology, Original draft preparation}

\author[1,2]{An Wang}[orcid=0000-0001-5515-0653]
\ead{wa09@link.cuhk.edu.hk}

\credit{Investigation, Original draft preparation}

\author
[1,2,4]{Hongliang Ren}[orcid=0000-0002-6488-1551]
\cormark[1]

\ead{hlren@ee.cuhk.edu.hk}
\credit{Conceptualization, Supervision, Review and editing, Resources}

\affiliation[4]{organization={Department of Biomedical Engineering, National University of Singapore},
    country={Singapore}}

\cortext[cor1]{Corresponding author: Hongliang Ren.}

\fntext[fn1]{Long Bai, Guankun Wang, and Mobarakol Islam are co-first authors.}

\begin{abstract}
Medical visual question answering (VQA) bridges the gap between visual information and clinical decision-making, enabling doctors to extract understanding from clinical images and videos. In particular, surgical VQA can enhance the interpretation of surgical data, aiding in accurate diagnoses, effective education, and clinical interventions. However, the inability of VQA models to visually indicate the regions of interest corresponding to the given questions results in incomplete comprehension of the surgical scene. To tackle this, we propose the surgical visual question localized-answering (VQLA) for precise and context-aware responses to specific queries regarding surgical images. Furthermore, to address the strong demand for safety in surgical scenarios and potential corruptions in image acquisition and transmission, we propose a novel approach called Calibrated Co-Attention Gated Vision-Language (C$^2$G-ViL) embedding to integrate and align multimodal information effectively. Additionally, we leverage the adversarial sample-based contrastive learning strategy to boost our performance and robustness. We also extend our EndoVis-18-VQLA and EndoVis-17-VQLA datasets to broaden the scope and application of our data. Extensive experiments on the aforementioned datasets demonstrate the remarkable performance and robustness of our solution.
Our solution can effectively combat real-world image corruption. Thus, our proposed approach can serve as an effective tool for assisting surgical education, patient care, and enhancing surgical outcomes. Our code and data will be released at~\hyperlink{https://github.com/longbai1006/Surgical-VQLAPlus}{github.com/longbai1006/Surgical-VQLAPlus}.
\end{abstract}




\maketitle

\section{Introduction}
\label{sec:1}

Comprehensive responses to inquiries on surgery and surgical procedures should be ideally provided by proficient surgeons. However, these scarce highly skilled surgeons often face difficulties in finding time to respond to such queries due to their demanding academic and clinical workload~\cite{seenivasan2022surgical,sharma2021medfusenet}. While recorded surgical videos have been provided to medical students to facilitate learning through observation, observational learning does not cater to specific queries raised by individual students. Its effectiveness largely depends on the student's ability to infer conclusions from videos. To address this to an extent, Surgical-Visual Question Answering (VQA)~\cite{seenivasan2022surgical} was introduced to answer questions on robotic surgery based on the given surgical scene. To help students better understand the predicted answer, we earlier introduced surgical Visual Question Localized-Answering (VQLA)~\cite{bai2023surgical}, which extended the VQA task further by localizing the regions in the image that are highly relevant to the predicted answer. Extending our earlier work, here, we propose VQLA++ to improve the performance and robustness of VQLA in robotic surgery. 

Numerous VQA models have emerged in the computer vision domain in recent times, showcasing impressive performance by effectively capturing semantic associations between images and questions~\cite{li2019visualbert,uppal2022multimodal,zhang2021multimodal,zhang2021dmrfnet}. Advanced techniques such as multimodal fusion~\cite{nguyen2023101868,bai2021influence} and attention mechanisms~\cite{zhang2020multimodal} have also been explored and incorporated to enhance performance. For their potential use in assisting medical professionals in analyzing medical images and providing insightful answers to specific queries, VQA models have gained significant attention in the medical domain~\cite{do2021multiple,khare2021mmbert}. A medical VQA system that addresses the unique challenges in the medical domain, such as complex anatomical structures, diverse imaging modalities, and domain-specific terminology, has the potential to serve as a reliable tool that can complement medical experts in responding to questions posed by patients and students~\cite{liu2023q2atransformer,seenivasan2022surgical,takada2020estimation}.

\begin{figure*}[]
    \centering
    \includegraphics[width=0.88\linewidth, trim=0 480 20 0]{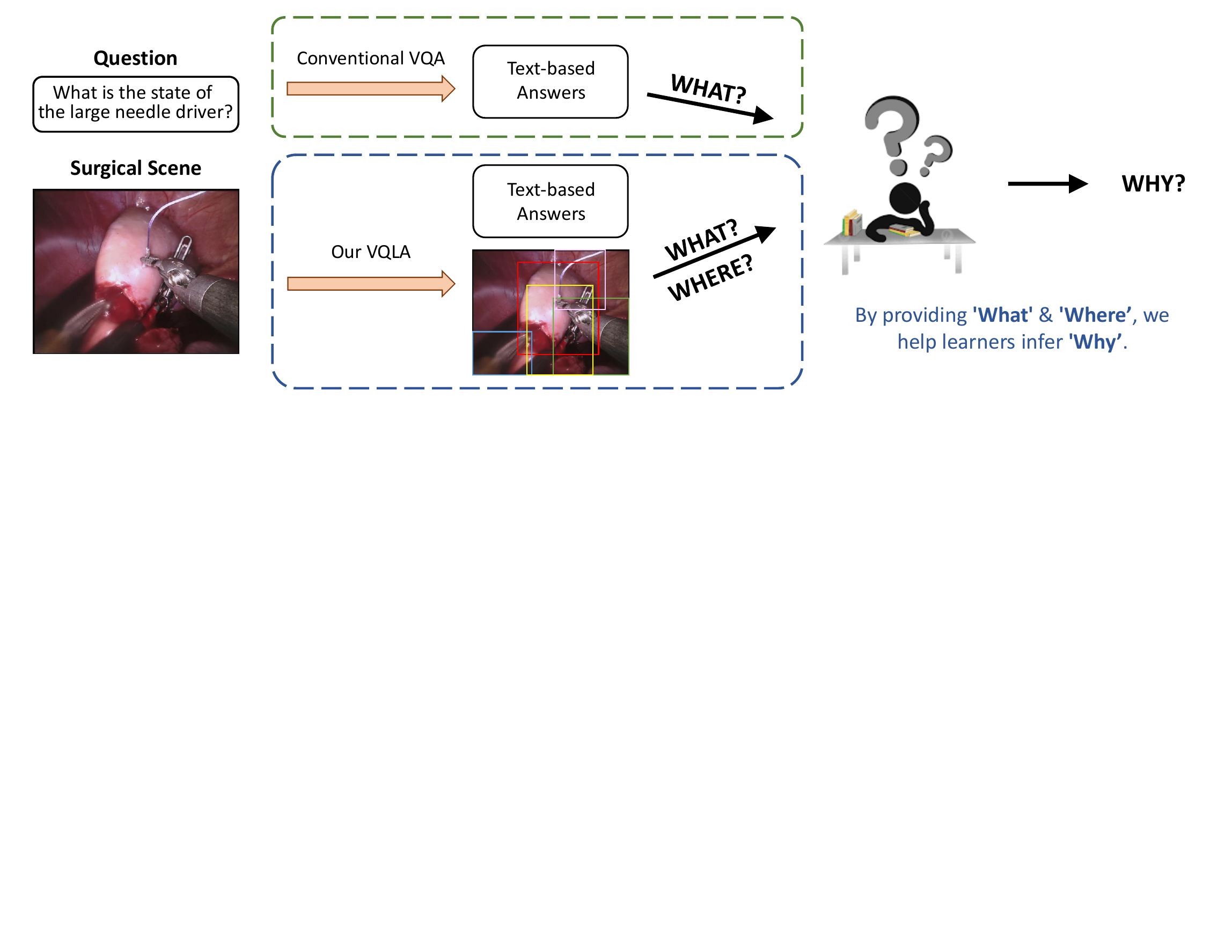}
    \caption{Comparison of the conventional VQA and our VQLA model. By providing the answer to 'What' and 'Where’, we can help learners to better infer 'Why’ and achieve a better understanding of surgical scenes.}
    \label{fig:overview}
\end{figure*}

Surgical-VQA~\cite{seenivasan2022surgical} has made a significant contribution to the medical surgical sub-domain by introducing VQA in robotic surgery. However, certain limitations hinder its effectiveness in assisting students with the comprehension of intricate surgical scenarios. In situations where expert guidance is unavailable for immediate responses to inquiries, understanding the rationale behind the answers often necessitates answers to both the "what" and "where" aspects. Traditional VQA approaches primarily focus on generating answers to the "what" question, neglecting the crucial dimension of "where" in the surgical scene the answer was derived from. This limitation poses challenges for students and trainees in deducing the reasons behind the predicted answer. For example, when a student queries the interaction between a surgical instrument and tissue, the VQA model can only provide a basic response without explicitly identifying the precise location of the tool and tissue within the surgical scene. Consequently, students still require assistance in comprehending the intricacies of complex surgical scenarios. Furthermore, the annotation of large amounts of textual data requires the involvement of expert surgeons, resulting in time-consuming and labor-intensive processes. To tackle these challenges, we earlier proposed the Surgical-VQLA~\cite{bai2023surgical}, which aims to highlight the specific area relevant to the posed question, as depicted in Figure~\ref{fig:overview}. By integrating the answer to the "what" question with the localization results and observations of the surgical area, students can achieve a more comprehensive understanding of specific surgical procedures, instrument usage, and anatomical structures involved. Additionally, observing the surrounding surgical area provides valuable contextual information that facilitates students' comprehension of the relationships and interactions among different elements in the surgical scene. By leveraging this integrated information, students can gain deeper insights into the reasoning behind surgical procedures. This enriched understanding enables them to infer the underlying causes and decisions, thereby enhancing their analytical capabilities and enabling more informed learning experiences.

\begin{figure}[t]
    \centering
    \includegraphics[width=\linewidth]{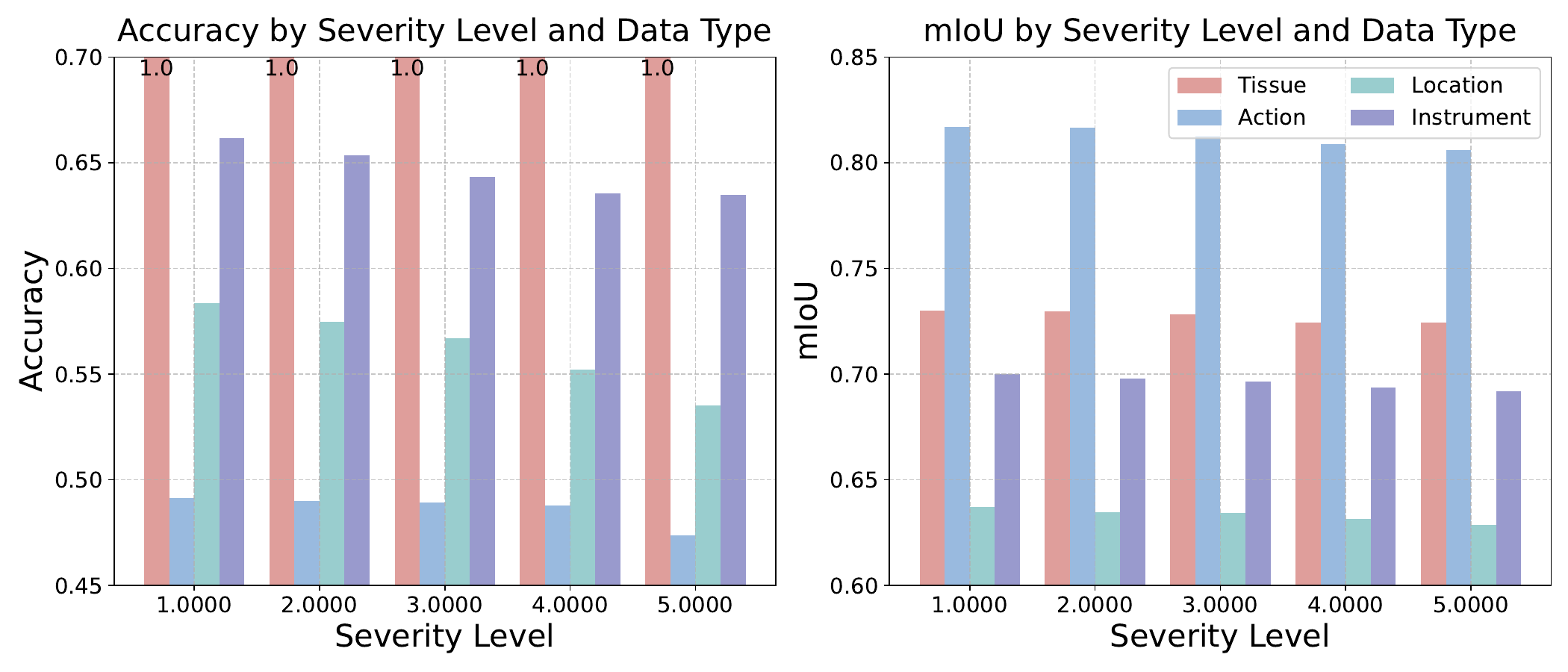}
    \caption{Robustness results of Surgical-VQLA~\cite{bai2023surgical} for different types of samples at each severity level. The four types of samples include the tissue being operated on, the current state of the instruments, the location of the instruments in the operating area, and the identification of the instruments. We can observe that as the severity level of corruption increases, the performance of the model significantly decreases.}
    \label{fig:motivation}
\end{figure}

In our earlier work~\cite{bai2023surgical}, apart from defining the Surgical-VQLA system, we addressed the challenges of detection-based visual feature extractor, to solve the problem of non-end-to-end, non-global understanding, and slow inference speed. We also effectively balanced the fusion between two modalities using a gated module. Moreover, we further proposed CAT-ViL~\cite{bai2023co}, which utilized a co-attention module to facilitate interaction between the two modalities, achieving improved fusion performance.
However, when applied in medical and clinical settings, models may face the challenge of maintaining high robustness. Medical data often contain various types of noise, artifacts, or image quality degradation. Figure~\ref{fig:motivation} shows the effect that different levels of image corruption cause on the performance of the existing model Surgical-VQLA~\cite{bai2023surgical}. There is a significant performance degradation as the level of image corruption increases. Moreover, these data are typically collected from different time points and institutions, leading to domain gaps~\cite{bai2023revisiting}. Therefore, it is imperative to develop a robust model capable of handling the sensitivity to diverse interfering factors and delivering stable and reliable performance in real-world scenarios. Robust models can handle various types of corruption efficiently and help minimize the risk of misdiagnosis, enabling healthcare professionals to make more accurate and informed decisions regarding patient care~\cite{islam2023robustness}. To address these challenges, we propose a joint framework that integrates feature calibration and adversarial contrastive training techniques. This integration aims to enhance the model's robustness and ensure its efficacy across various practical applications.

By introducing multimodal collaborated calibration and global contextual calibration mechanisms, we develop a new Surgical-VQLA++ framework. This model effectively aligns and normalizes multimodal representations and enhances robustness by capturing subtle feature perturbations through contrastive representation learning of adversarial examples. Through extensive experiments, our proposed solution has demonstrated significant improvements in performance and robustness. Our model surpasses the second-best model by achieving an improvement of 1.12\% in accuracy and 1.55\% in mIoU in overall performance. In robustness tests, it further exceeds the second-best by 1.64\% in accuracy and 1.70\% in mIoU, demonstrating the effectiveness of our proposed method. 

In this work, our key contributions are:
\begin{itemize}
    \item [--] We present a Surgical-VQLA++ framework that establishes an instance-level connection between answering and localization. The proposed model exhibits impressive performance and robustness, rendering it highly suitable for clinical applications. Furthermore, our feature extraction strategy promotes global understanding, compatibility with end-to-end solutions, and an efficient inference speed of 150.6 FPS.
    \item [--] To enhance the alignment and interaction of multimodal representations, as well as globally calibrate contextual features, we propose a \textbf{C}alibrated \textbf{C}o-attention \textbf{G}ated \textbf{Vi}sion-\textbf{L}anguage (C$^2$G-ViL) embedding module. This module explores optimal fusion weights and facilitates the alignment between different modalities.
    \item [--] Building upon the DeiT backbone~\cite{touvron2021training} for deep feature learning, we employ an adversarial contrastive training strategy specifically on the DeiT backbone. This strategic utilization aims to enhance the model's capacity to discern and capture intricate nuances or patterns, further boosting model performance and robustness. For multi-task convergence, we investigate various combinations of loss functions and move away from naive loss combinations.
    \item [--] We present two comprehensive surgical datasets for VQLA tasks based on the public EndoVis18~\cite{allan2020endovis18} and EndoVis17~\cite{allan2019endovis17} datasets. We have expanded our dataset to include a total of 17,269 QA pairs by adding additional queries about surgical instruments. These QA pairs cover contents such as surgical organs, instruments, actions, and instrument locations. Each pair also comes with a corresponding bounding box to localize the answer.
\end{itemize}

\section{Related Work}
\label{sec:2}
\subsection{Grounded VQA in Computer Vision}

VQA and answer grounding in computer vision have gained attention in recent times~\cite{manmadhan2020visual}. Antol \textit{et al.}~\cite{antol2015vqa} introduced the VQA task to answer natural language questions based on the input images. Subsequently, numerous multimodal models incorporating advanced techniques were introduced in the computer vision domain to improve the performance on VQA tasks~\cite{le2023guiding,li2019visualbert,yu2019deep}. The potential use of the VQA model for critical tasks has emphasized the exploration of model interpretability~\cite{parelli2023interpretable}. Visual grounding, an emerging topic within the vision-language tasks~\cite{deng2018visual,su2023language, jiang2022pseudo}, is being widely explored in the development of interpretable models and dependable scene understanding in VQA. Incorporating relevant visual evidence through language queries has been observed to help effectively guide the model's attention and contribute to achieving state-of-the-art (SOTA) performance in VQA tasks~\cite{urooj2021found}. As a result, the effective execution of both question-answering and answer grounding has emerged as a prominent research area.

The integration of VQA and answer grounding techniques facilitates a more profound comprehension of visual content in computer vision research. Through the grounding process, answers can be linked to specific regions or objects within an image, providing precise localization information. This integration not only enables accurate answer generation but also offers explanations by connecting the selected answer to relevant visual evidence, thereby enhancing interpretability~\cite{kv2020reducing}. Achieving interpretability is particularly important in critical domains like healthcare, where building a trustworthy and robust understanding of computer vision systems is essential. Numerous studies have focused on developing advanced deep learning models integrating VQA and answer grounding. For instance, Fukui \textit{et al.}~\cite{fukui2016multimodal} proposed a compact bilinear pooling model for VQA and visual grounding. In contrast, Zhu \textit{et al.}~\cite{zhu2023dual} introduced a dual-decoder transformer network specifically designed for answer grounding in VQA. Zhang \textit{et al.}~\cite{zhang2019interpretable} proposed an interpretable VQA approach by incorporating visual grounding through attention supervision mining. Reich \textit{et al.}~\cite{reich2023measuring} presented a method to measure the faithfulness and plausibility of visual grounding in VQA. By guiding the model's attention to the most salient areas of an image relevant to the question, answer grounding enhances reasoning abilities and improves answer accuracy. Mani \textit{et al.}~\cite{mani2020point} integrated object proposals into VQA by utilizing pointing to indicate the location of the answer within the image. However, these approaches, designed for natural images, pose challenges in terms of inference speed and global contextual understanding due to their object proposal design. The medical domain, with its complex and sensitive medical imaging data, presents unique challenges. Adapting VQA and answering grounding tasks to the medical domain requires addressing concerns such as interpretability, transparency, and uncertainty, making it an active area of research.

\subsection{Grounded VQA in the Medical Domain}
Medical VQA has made significant contributions to the medical field, utilizing the fusion of images and questions to assist with clinical diagnosis and treatment. This includes aiding radiologists~\cite{liu2021contrastive}, enhancing clinical decision support~\cite{naseem2023k}, facilitating patient monitoring and assessment~\cite{seenivasan2022surgical}, supporting education and training~\cite{seenivasan2023surgicalgpt,wang2024surgical}, enabling clinical research and data analysis~\cite{lau2018dataset,chen2024asiseg}, and more. However, medical images possess unique characteristics that require specialized knowledge for accurate interpretation~\cite{bai2023llcaps,chen2024lightdiff,wang2023rethinking}, posing additional challenges in developing medical VQA models. Recent advancements in medical VQA research have focused on addressing these challenges. Liu \textit{et al.}~\cite{liu2021contrastive} proposed contrastive pre-training and representation distillation, Naseem \textit{et al.}~\cite{naseem2023k} introduced knowledge-aware multimodal representation, and Liu \textit{et al.}~\cite{liu2023q2atransformer} presented an answer querying decoder. These works have significantly improved performance in their respective tasks.

Despite these advancements, answer grounding or VQLA in the medical domain remains relatively unexplored. Tascon \textit{et al.}~\cite{tascon2022consistency} discussed localized questions but cropped the target regions, which limits global understanding. They later improved their solution by localizing the question in the image using attention maps to enhance VQA performance~\cite{tascon2023localized}. Similarly, Vu \textit{et al.}~\cite{vu2020question} employed multi-glimpse attention to localize image regions relevant to given questions. However, these methods did not consider the framework of multi-task learning, focusing only on how to improve the performance of VQA itself by localizing answers. Instead, the VQLA system can simultaneously output question-answers and localizations, offering several potential benefits. Another interesting study locates relevant video snippets from medical instructive videos to answer textual questions~\cite{li2022towards}. However, this study is not directly applicable to the tasks of VQA or VQLA as it solely focuses on locating matching spans in videos that are relevant to given questions. 

The multi-task learning framework can leverage shared representations and interactions between question-answering and localization, improving performance in both tasks via joint training and information sharing~\cite{cai2023multi,zhang2021multi}. Additionally, simultaneously outputting question-answering and localization can provide more comprehensive and informative results. The model can offer visual explanations or justifications for its answers, enhancing the interpretability and trustworthiness of the system.

\section{Methodology}
\label{sec:3}
Our VQLA system architecture, as illustrated in Figure~\ref{fig:network}, incorporates two feature extractors, namely the C$^2$G-ViL embedding module and the pre-trained DeiT backbone. Additionally, it includes two prediction heads responsible for classification and bounding box generation tasks. To optimize the model's performance, we employ the adversarial contrastive training strategy during the training process.
\subsection{Preliminaries}
\label{sec:preliminaries}
\subsubsection{VisualBERT}
\label{sec:visualbert}
VisualBERT~\cite{li2019visualbert} performs joint reasoning on vision and language tasks, by integrating Transformer~\cite{vaswani2017attention} models and Faster RCNN~\cite{ren2015faster}, which extended the BERT~\cite{devlin2018bert} NLP model. Initially, tokens were generated from textual inputs, while visual features were extracted using Faster RCNN~\cite{ren2015faster}. VisualBERT then employed element-wise addition operations on the token, segment, and positional embedding from the textual input, in addition to the feature, segment, and positional embedding from the visual input. These operations allowed both modalities to be combined effectively through concatenation. Subsequently, Transformer layers were utilized to boost interactions between the two modalities, which facilitated the learning of joint representations. Finally, prediction modules specific to various ViL tasks, such as VQA, visual answer justification, and visual reasoning were designed to perform predictions. The simplistic, efficient, and prompt design of multimodal fusion in VisualBERT effectively enhances subsequent modules to facilitate multimodal joint reasoning.

\subsubsection{Multi-head Attention}
\label{sec:mha}

Multi-head attention (MHA)~\cite{vaswani2017attention} is a mechanism that facilitates simultaneous attention to multiple regions of the input, enabling the model to capture diverse and complementary information effectively. Unlike relying solely on a single attention mechanism, multi-head attention involves the creation of multiple sets of attention weights, referred to as heads, which are computed independently. Each head attends to different portions of the input, thereby providing multiple perspectives and insights into the underlying data. The outputs generated by these heads are subsequently combined using linear transformations, thereby enhancing the model's capacity to capture intricate relationships and improve overall performance. In our formulation, we represent the input queries, keys, and values for each head $i$ as $Q_i$, $K_i$, and $V_i$, respectively, where $i$ ranges from $1$ to $h$. Accordingly, let $Attn$ denote the single-head attention mechanism, and consequently, we can express this as follows:
\begin{equation}
    h_i=Attn\left(\boldsymbol{W}_{Qi} \boldsymbol{Q}_i, \boldsymbol{W}_{Ki} \boldsymbol{K}_i, \boldsymbol{W}_{Vi} \boldsymbol{V}_i\right)
\end{equation}

A linear transformation is subsequently utilized for the aggregation of attention across multiple heads. $\bm{W}_o$ represents the trainable parameters in the multiple heads, and the MHA shall be formulated as:
\begin{equation}
h=MultiAttn \; (\boldsymbol{W}_o\left[h_1 \| \dots \| h_h \right])
\end{equation}
in which $[\cdot\,\|\,\cdot]$ represents concatenation.
\begin{figure*}[t]
    \centering
    \includegraphics[width=\linewidth, trim=5 240 68 0]{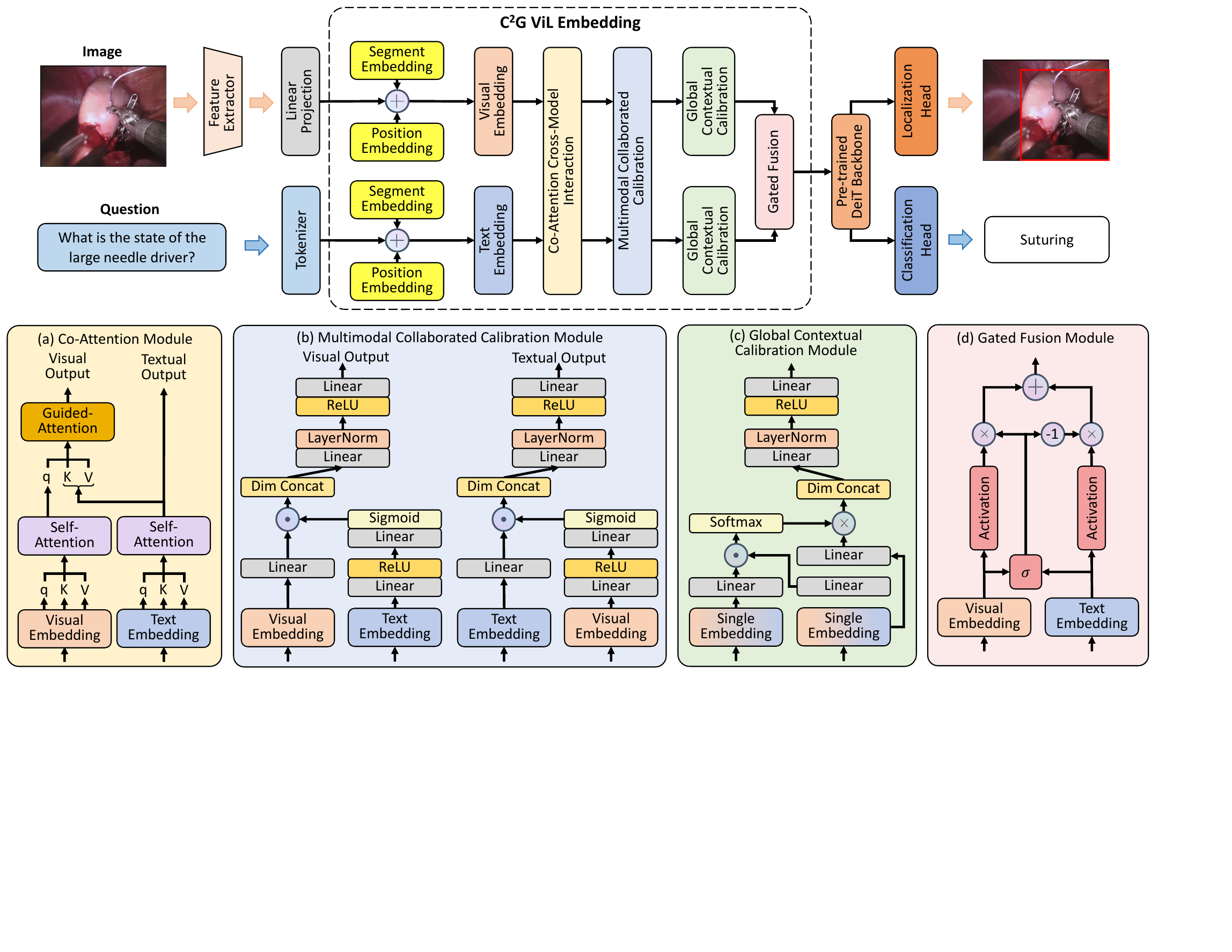}
    \caption{The overall network architecture of our Surgical-VQLA++ framework. The network comprises a visual feature extractor, a customizedly trained tokenizer, C$^2$G-ViL embedding module (embedding setup, co-attention cross-model interaction, multimodal collaborated calibration, global contextual calibration, gated fusion), per-trained DeiT backbone, and the parallel prediction heads for question answering and answer localization. $\odot$, $\otimes$, and $\oplus$ represent the element-wise dot product, cross product, and summation operation, respectively. Dim Concat denotes dimensional concatenation.}
    \label{fig:network}
\end{figure*}

\subsubsection{Adversarial Examples}
\label{sec:at}
Adversarial examples are carefully constructed inputs that exploit vulnerabilities in the decision-making process of machine learning models to deceive them. These examples are specifically designed to closely resemble the original data while incorporating subtle modifications, known as perturbations. Perturbations refer to minor alterations introduced to the input data, often imperceptible to human observers, with the objective of inducing incorrect predictions or unexpected behavior from the model.
Goodfellow~\emph{et al.}~\cite{goodfellow2014explaining} proposed Fast Gradient Sign Method (FGSM) to generate adversarial examples. We follow~\cite{pan2022improved} to perform perturbation on the fused embedding. Let $x$ and $G$ denote the input data and ground truth, respectively. Given a deep model $F$, the loss function can be defined as $\mathcal{L}(F(x+r),G)$, where $r$ represents the perturbation. With enforcing a maximum constraint on $r$, we can obtain $ \{ \max _r \mathcal{L}(F(x+r),G) \} \; s.t. \; \|r\|_{\infty}<\epsilon$.
It can be approximated via the first-order approximation:
\begin{equation}
    \mathcal{L}(F(x+r),G) \approx \mathcal{L}(F(x),G)+\nabla_{x} \mathcal{L}(F(x),G)^T r
\end{equation}
Finally, we can obtain the perturbation as:
\begin{equation}
    r=-\epsilon \operatorname{sign}\left(\nabla_{x} \mathcal{L}(F(x),G)\right)
\label{equ:adv}
\end{equation}

\subsection{Feature Extraction}
\label{sec:feature_extraction}
Two parallel feature extractors are employed in this study to handle visual and textual representations separately. For visual representation, a visual feature extractor is utilized to obtain feature representations from original images. However, unlike the original VisualBERT~\cite{li2019visualbert}, instead of adopting an object proposal-based strategy, we employ a ResNet18~\cite{he2016deep} network that has been pre-trained on ImageNet~\cite{deng2009imagenet}. This choice is made because object detection models solely focus on key objects and fail to consider surgical instruments and the surrounding environment, which play a crucial role in prediction accuracy~\cite{seenivasan2022global}. Moreover, any errors in the detection process could significantly impact the final prediction results. In contrast, ResNet18 serves effectively as a global feature extractor, providing a holistic understanding of the target image compared to Faster RCNN~\cite{ren2015faster}, which is limited to local key objects. Additionally, pre-trained ResNet18 offers faster inference speed and an end-to-end solution.

In parallel to the visual feature extractor, a tokenizer is applied, specifically trained on the surgical dataset, to convert input questions into embedding matrices, which can subsequently be processed by the deep model.
Following this, features from both modalities are fed into the C$^2$G-ViL embedding module for further processing and fusion.

\subsection{C$^2$G-ViL Embedding}
\label{sec:c2gvil}

In this study, we initially employ the VisualBERT strategy to obtain separate visual and text embedding. However, previous approaches have taken a simplistic approach to data fusion, utilizing concatenation~\cite{li2019visualbert,seenivasan2022surgical}, summation~\cite{wu2022two}, or MLP layer~\cite{yu2019deep}. This rudimentary fusion method fails to capture the active interaction between information and statistical representations from the two modalities, thereby limiting their effectiveness in subsequent tasks. To address this limitation, we propose a novel Calibrated Co-Attention Gated (C$^2$G)-ViL embedding that aims to calibrate and aggregate the visual and textual representations.

Our C$^2$G-ViL embedding comprises a co-attention interaction module, a multimodal collaborated calibration module, a global contextual calibration module, and a gated fusion module.
Firstly, the co-attention module integrates guided-attention and self-attention mechanisms to enhance the interaction between modalities. By incorporating guided-attention blocks, our model is able to direct its attention towards key regions in images, further improving its understanding of visual content.
Secondly, two calibration modules are engaged to ensure the robustness of models for reliable performance in clinical settings. These models need to accommodate variations in data, including diverse patient populations, imaging techniques, and image quality. Therefore, we introduce a process of representation calibration to enhance performance by reducing variations and discrepancies within the embedding space. This calibration process facilitates more accurate feature differentiation and dependable predictions~\cite{yuan2022detecting}.
Finally, we leverage a gate fusion operation to dynamically learn and control the fusion weights of different modalities. This dynamic fusion approach aims to achieve a more effective integration of multimodal representations, optimizing the overall performance of the model.

\subsubsection{Co-Attention Cross-Model Interaction}
The co-attention block utilized in this study aims to capture the relationships and dependencies between two inputs. Motivated by the works~\cite{chen2021crossvit, yu2019deep}, we introduce a co-attention learning block to enhance cross-modality interaction, facilitating information exchange and leveraging the complementary features and contextual cues present in the multimodal inputs~\cite{bai2023co}. Each co-attention layer incorporates a self-attention block for each modality to learn contextually rich representations. Moreover, we employ a guided-attention block to guide the visual input based on the text input, enabling effective identification and focus on target image regions. The MHA block~\cite{vaswani2017attention} described in Section~\ref{sec:mha} is a typical example. The guided-attention and self-attention blocks share the same architecture, although their inputs differ. Specifically, to allow the visual input to attend to relevant information from the text input, the guided-attention block takes queries from the visual input and keys and values from the text input:
\begin{equation}
    h_i=Attn\left(\boldsymbol{W}_{Qi} \boldsymbol{Q}_i^{\rm Visual}, \boldsymbol{W}_{Ki} \boldsymbol{K}_i^{\rm Text}, \boldsymbol{W}_{Vi} \boldsymbol{V}_i^{\rm Text}\right)
\end{equation}
The network structure of the co-attention module is depicted in Figure~\ref{fig:network}~(a). Through this mutual interaction, the model acquires a comprehensive understanding of the instructive interplay from textual input to visual input, resulting in improved performance in joint reasoning. Empirically, we employ six co-attention layers to establish the interactive relationship between the two modalities. The co-attention mechanism effectively enables the model to harness the synergistic effects between the inputs, enhancing the overall representation and facilitating more accurate and informed decision-making.

\subsubsection{Multimodal Collaborated Calibration}
The multimodal collaborated calibration consists of two identical modules, in which one calibrates the visual embedding using text embedding, and the other calibrates the text embedding using visual embedding, as shown in Figure~\ref{fig:network}~(b). This module aims to align the representation from different modalities by establishing a reliable correspondence between them.

To calibrate the visual embedding based on the text embedding, as depicted in Figure~\ref{fig:network}~(b), the two input embeddings are first projected into $N$ calibration heads by the first two linear layers, where, each head corresponds to embedding dimension $\frac{D}{N}$. The text embedding is then propagated through a linear layer with Sigmoid activation function. This operation shall effectively align the text embedding with the visual embedding. The output representation can be adjust according to the attention weights. Then, the element-wise product is conducted between the visual and text embedding, before the concatenation of multiple calibration heads. The subsequent fully connected layer, LayerNorm, and ReLU activation will further shape the output of the calibrated embedding.

Calibrating text embedding using visual inputs follows the same steps, with the only difference being the swap of the two modalities. Therefore, we can utilize the mutual statistical correlation to quantify the relationship between patterns. The resulting calibrated embedding shall capture the alignment and refinement information from both visual and text modalities.

\subsubsection{Global Contextual Calibration}
Following the calibration of the joint embedding, the module depicted in Figure~\ref{fig:network}~(c) aims to refine and enhance the global contextual semantics within each modality. To achieve this, we utilize the pairwise bilinear pooling technique to refine and explore the contextual information~\cite{kim2016hadamard}. The global contextual calibration module comprises two structurally identical components, with each component operating on the visual and textual embedding, respectively. We denote the input embedding from the same modality as $E_n$ and $E_m$. They are directly divided into $N$ calibration heads with embedding dimension $\frac{D}{N}$ in each head.
\begin{equation}
\begin{split}
    & E_{m}=\left\{E_{t,i} \in \mathbb{R}^{\frac{D}{N}}\right\}_{i=1}^{N} \\
    & E_{n}=\left\{E_{v,i} \in \mathbb{R}^{\frac{D}{N}}\right\}_{i=1}^{N}
\end{split}
\end{equation}
The formulation of the bilinear pooling can be expressed as follows:
\begin{equation}
    E_{gcc}= \mathbf{Softmax} \left[{\theta}^{\mathrm{T}}{L}_{1} \left({E}_n\right) \odot {L}_{2} \left({E}_m\right)\right] \times {L}_{3} \left({E}_m\right)
\end{equation}
in which $\theta$ is the learnable weight vector, and $L$ denotes the linear layer. $L_{1}, L_{2}\in \mathbb{R}^{\frac{D}{N} \times \frac{D}{N}}$, and $L_3 \in \mathbb{R}^{\frac{D}{N} \times D}$.
Then the projected embedding from different heads will be concatenated to obtain an updated embeddding $E_{gcc}'$.
The final output embedding shall be shaped through:
\begin{equation}
    E_{gcc}'' = E_n+{L}_{5}\left[\mathbf{ReLU}\left[\mathbf{LN}\left({L}_{4}\left(E'_{gcc}\right)\right)\right]\right]
\end{equation}
in which $L_{4}, L_{5} \in \mathbb{R}^{D \times D}$, and $\mathbf{LN}$ denotes LayerNorm. Consequently, we perform context calibration on each modality separately based on the statistical information, improving global reasoning capabilities.

\subsubsection{Gated Fusion}
A mere concatenation or linear combination of multiple modality inputs lacks the ability to effectively facilitate the learning of cross-modality representations~\cite{arevalo2017gmu}. To address this limitation, we have introduced a gating mechanism to regulate the inputs from different modalities following our preliminary conference version~\cite{bai2023surgical}. The specific node $\delta$ dynamically adjusts the combination weights for the visual and textual embedding inputs in the subsequent step, allowing the gating module to learn the fusion representation and weights directly from the training data. The $tanh$ activation function is employed to encode each modality individually. Let ${R}_v$ and ${R}_t$ denote the visual and text inputs, respectively. The formulation of the gated module can be expressed as follows:
\begin{equation}
    \begin{aligned}  
       & {I} = k * \tanh \left(\omega_v\cdot {I}_v\right) +(1-k) * \tanh \left(\omega_t \cdot {I}_t\right) \\ 
       & {k} = \delta\left(\omega \cdot \left[{I}_v\| {I}_t\right]\right) \\ 
    \end{aligned}
    \label{equ:1}
\end{equation}
in which $\omega, \omega_v, \omega_t$ are learnable parameters. As presented in Figure~\ref{fig:network}~(d), this differential solution effectively couples the multimodal embedding inputs, aiming to find a better intermediate representation for the combination of the two modalities.  

\subsection{Contrastive Training with Adversarial Examples}
\label{sec:acl}

\begin{figure*}[]
    \centering
    \includegraphics[width=\linewidth, trim=10 670 160 0]{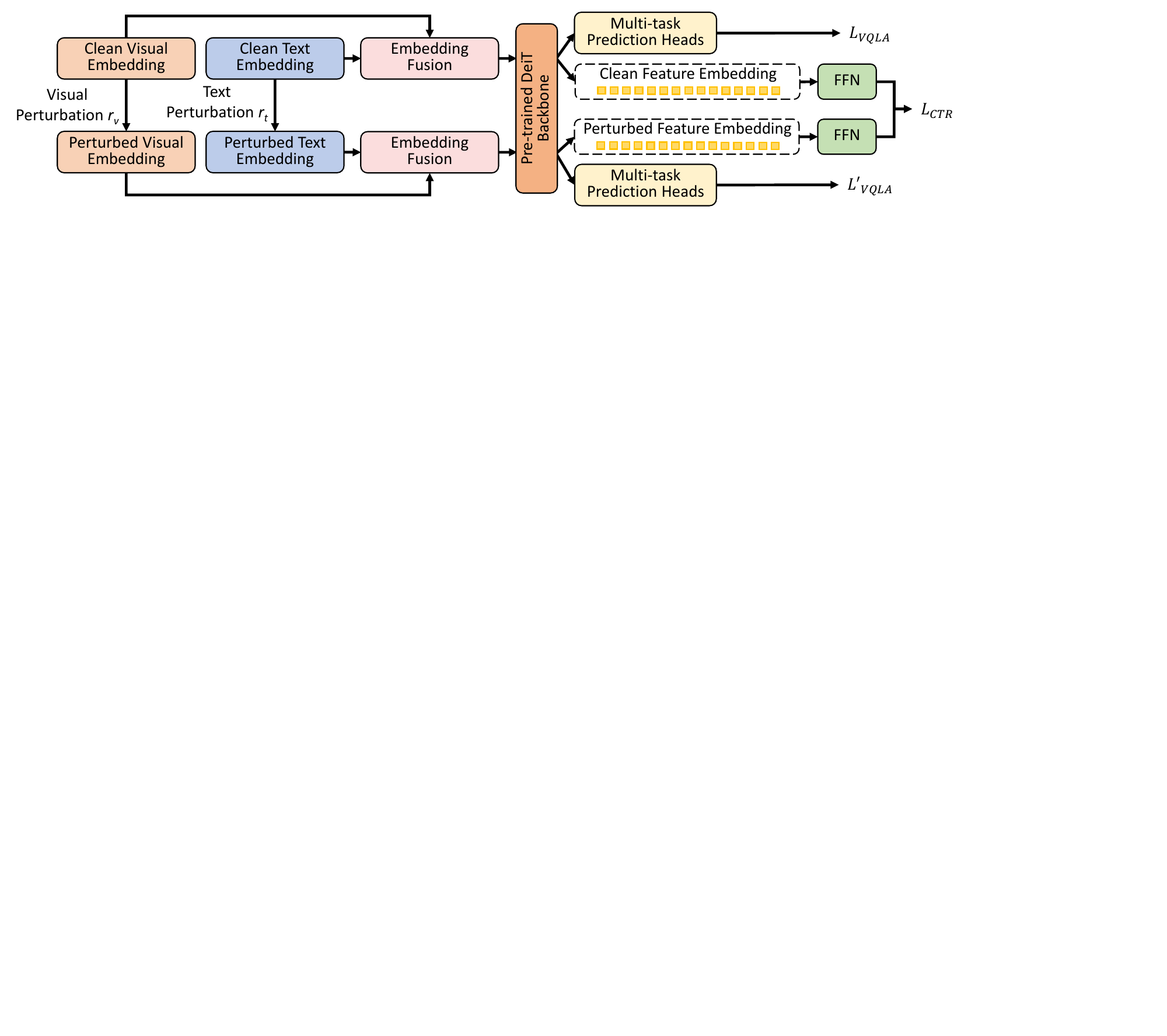}
    \caption{Overview of the adversarial contrastive training strategy. The visual and text embedding are perturbed separately. After being propagated through the embedding fusion and pre-trained DeiT backbone, the contrastive loss is applied to the clean and perturbed feature embedding.}
    \label{fig:acl}
\end{figure*}

Despite their intent to deceive machine learning models, adversarial examples can paradoxically enhance the robustness of the model through the adversarial training process. Exposing the model to adversarial examples during training can force it to learn more generalized and robust representations resilient to small perturbations in the input space. Therefore, as presented in Figure~\ref{fig:acl}, we utilize the adversarial contrastive training strategy to augment the learning process, which encourages the model to consider a broader range of potential inputs and variations, thereby enhancing the robustness and generalization capabilities.

Specifically, we compute the gradients of text and visual embedding separately, and apply the adversarial perturbation described in Equ.~(\ref{equ:adv}). $r_t$ and $r_v$ denote the perturbations applied on text and visual embedding, respectively:
\begin{equation}
\begin{split}
    & r_t=-\alpha \cdot \epsilon \operatorname{sign}\left(\nabla_{x_t} \mathcal{L}(F(x_t),G)\right) \\
    & r_v=-\beta \cdot \epsilon \operatorname{sign}\left(\nabla_{x_v} \mathcal{L}(F(x_v),G)\right)
\end{split}    
\label{equ:adv_tv}
\end{equation}
in which $\alpha$ and $\beta$ represent the perturbation weight on each modality. Subsequently, after the embedding fusion, the clean and perturbed embedding shall be propagated through the backbone network for deep feature learning. Following~\cite{bai2023co}, we adopt the DeiT-base~\cite{touvron2021training} pre-trained on ImageNet~\cite{deng2009imagenet} as the backbone.

To facilitate efficient representation learning, we incorporate a regularization approach based on contrastive learning, leveraging both clean and perturbed embedding examples. This regularization technique aims to promote the learning of robust representations that encompass both intraclass similarities and interclass differences. The clean and perturbed example pairs are propagated through both the pre-trained DeiT backbone and a feed-forward network (FFN) to obtain the clean feature matrix $P_i$ and perturbed feature matrix $P_j^r$.
The cosine similarity is employed to get the distance between $P_i$ and $P_j^r$:
\begin{equation}
    \mathbf{cos}\left(P_i, P_j^r \right) = \frac{(P_i) (P_j^r)^{\mathrm{T}}}{(P_i)^2 \cdot {(P_j^r)^2}}
\end{equation}
According to~\cite{chen2020simple}, given a batch of $B$ clean examples, we generate $B$ adversarial examples, resulting in a total of $2B$ examples. 
The contrastive loss is thereby established as follows:
\begin{equation}
\mathcal{L}_{CTR} (x;r)=- \log \frac{\exp \left( {\mathbf{cos}\left({P_i}, \; {P_j^r}\right) /C }\right)}{\sum_{\gamma=1}^{2 B}  \| \exp \left(\mathbf{cos}\left({P_i}, \; {P_\gamma^r}\right) /C \right) \|_{0 | \gamma = i}}
\end{equation}
in which $C$ represents the regularization temperature. $\|\cdot\|_{0 | \gamma = i}$ means the result is 0 when $\gamma = i$, otherwise it remains unchanged. By incorporating both clean feature examples and adversarial examples within a contrastive learning framework, the model is able to achieve strong generalization capabilities that extend beyond the training data. This integration enables the development of robust and disentangled feature representations that effectively capture the intrinsic characteristics of the data. Furthermore, it enhances the model's robustness and resilience to various sources of variation. 

\subsection{Prediction Heads and Loss Functions:}
\label{sec:loss}
Following our preliminary conference version~\cite{bai2023surgical}, the parallel prediction heads comprise a linear layer for classification and a 3-layer FFN for bounding box regression. Besides, our preliminary version~\cite{bai2023surgical} employed the CE loss as our QA loss, and the summation of $\mathcal{L}_1$-norm and GIoU loss~\cite{rezatofighi2019generalized} as our localization loss, as follows:
\begin{equation}
\mathcal{L}'_{VQLA}(x;G) = \mathcal{L}_{CE} + \left(\mathcal{L}_{GIoU} +\mathcal{L}_{1}\right)
\end{equation}
However, in the context of question-answering (QA), there exists a significant challenge of class imbalance that prevents effective convergence when using the cross-entropy (CE) loss. To address this issue, we apply the focal loss~\cite{lin2017focal} to tackle the imbalanced QA prediction. In the case of localization, Carion \textit{et al.}~\cite{carion2020end} previously utilized a combination of $\mathcal{L}_1$-norm and GIoU loss for object detection. However, our extensive comparative experiments have shown that the use of only the GIoU loss can achieve even better performance.

\begin{figure*}[t]
    \centering
    \includegraphics[width=\linewidth, trim=80 535 85 0]{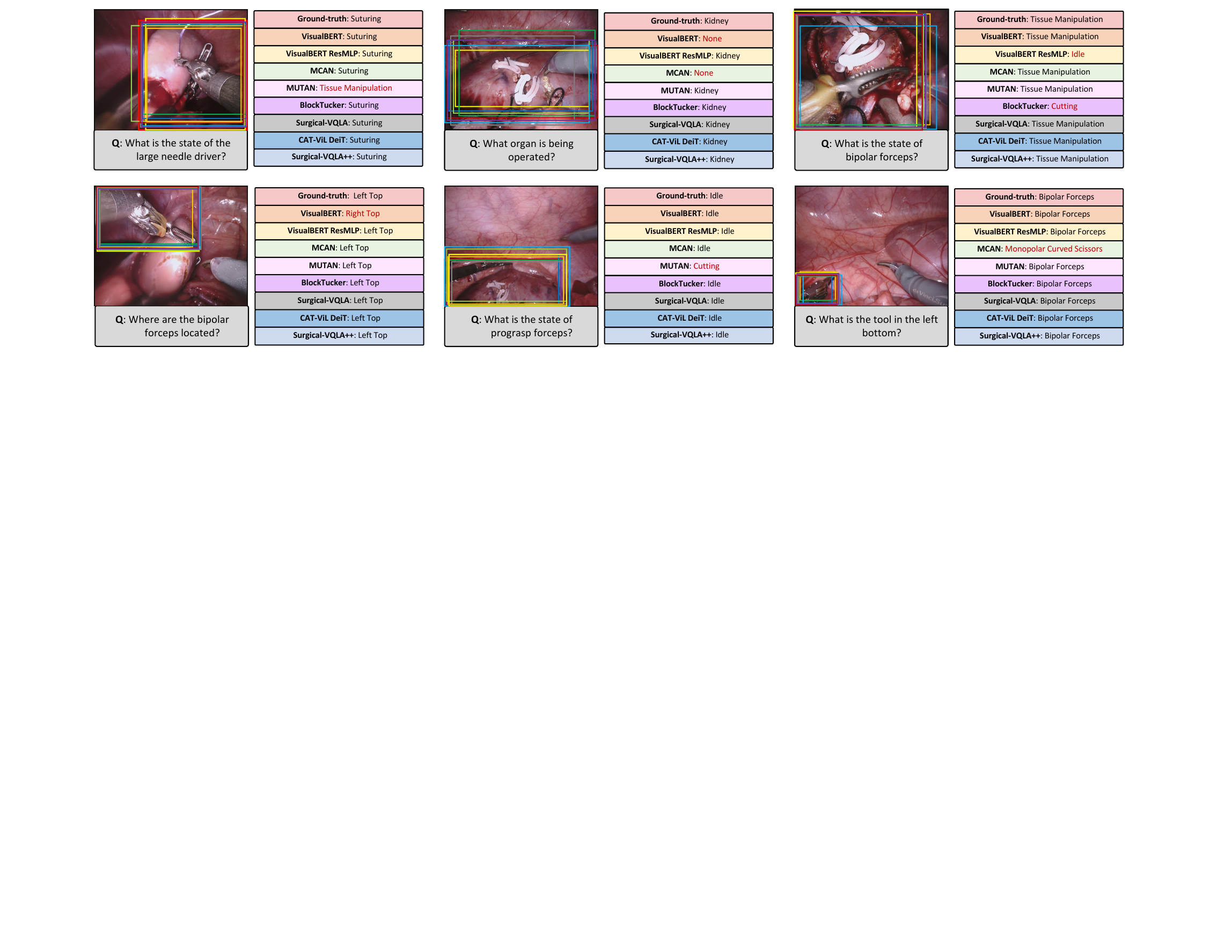}
    \caption{Qualitative comparison of our Surgical-VQLA++ against SOTA solutions. The color of the detection bounding box corresponds to the color of the answers, in which red represents the ground truth and light blue represents our proposed solution. On the four types of question-answer pairs (tissue, action, location, instrument), our method achieves the best performance on both answering and localization tasks.}
    \label{fig:visual}
\end{figure*}

Moreover, to effectively balance the weights between the two losses and address the trade-off inherent in multi-task learning, it is crucial to explore and determine the optimal weight parameters. 
Rather than performing this exploration manually, we leverage the uncertainty loss~\cite{kendall2018multi} for multi-task training. By treating dropout as a Bayesian approximation, the uncertainty loss shall sample multiple dropout masks. This process leads to multiple predictions for a given input, enabling the estimation of task uncertainties. These uncertainties are then used to weigh the losses during training.  
Tasks with lower uncertainties shall be assigned higher weights and vice versa. Therefore, the model can dynamically allocate weights and resources to the localization and question-answering tasks during training.
The resulting VQLA loss can be formulated as follows:
\begin{equation}
    \mathcal{L}_{VQLA}(x;G) = \frac{1}{2 \sigma_1^2} \mathcal{L}_{Focal} +\log \sigma_1 +\frac{1}{2 \sigma_2^2} \mathcal{L}_{GIoU}+\log \sigma_2
\end{equation}
where the scale value $\sigma_1$ and $\sigma_2$ are learnable parameters. Therefore, our final loss is the summation of VQLA loss on clean examples ($\mathcal{L}_{VQLA}$), VQLA loss on perturbed examples ($\mathcal{L}'_{VQLA}$), and the contrastive loss ($\mathcal{L}_{CTR}$):
\begin{equation}
    \mathcal{L} = \mathcal{L}_{VQLA}(x;G) + \mathcal{L}_{VQLA}(x+r;G) + \mathcal{L}_{CTR}(x;r)
\end{equation}

\section{Experiments}
\label{sec:4}

\subsection{Datasets}
\label{sec:datasets}

\noindent \textbf{EndoVis-18-VQLA} is a publicly available robotic surgery dataset derived from the MICCAI Challenge 2018~\cite{allan2020endovis18}. The VQLA labels for this dataset are accessible through~\cite{bai2023surgical}. Initially, the annotations only included surgical actions, target tissues, and instrument positions. To enhance the dataset's utility, we have expanded it by adding questions specifically targeting instrument names. This expansion aims to improve the overall usefulness and diversity of the dataset. The training comprises 1560 frames, and the test set has 436 frames. The number of QA pairs in the training set are expanded from 9014 to 12741, the test set is extended from 2769 to 3820 QA pairs. The dataset splitting follows the guidelines (Training Video ID: 2, 3, 4, 6, 7, 9, 10, 11, 12, 14, 15; Test Video ID: 1, 5, 16) in~\cite{seenivasan2022surgical} to ensure proper isolation between video sequences and prevent any potential data leakage.

\noindent \textbf{EndoVis-17-VQLA}, derived from the MICCAI Challenge 2017~\cite{allan2019endovis17}, is used for external validation. It is important to note that we \textit{train} our model solely on the EndoVis-18 training set and then \textit{test} its performance on the EndoVis-18 test set and EndoVis-17 dataset. The annotations for this dataset are also provided by~\cite{bai2023surgical}, and we apply the same dataset expansion as mentioned above. With 97 frames, this dataset is extended from 472 to 708 QA pairs. The purpose of using this dataset is to showcase the generalizability of our model when facing domain gaps. 

\begin{table*}[t]
\caption{Comparison experiments between the preliminary version Surgical-VQLA~\cite{bai2023surgical} and the SOTA version Surgical-VQLA++.}
\centering
\label{tab:ab1}  
\resizebox{0.85\textwidth}{!}{
\begin{tabular}{c|ccc|ccc|ccc}
\toprule[1pt]
\multirow{2}{*}{Models} & \multirow{2}{*}{Localization} & \multirow{2}{*}{Classification} & \multirow{2}{*}{Uncertainty} & \multicolumn{3}{c|}{EndoVis-18-VQLA}                       & \multicolumn{3}{c}{EndoVis-17-VQLA}                       \\ \cline{5-10}
                                &                                 &                              & &Acc               & F-Score           & mIoU              & Acc               & F-Score           & mIoU              \\ \hline

Surgical-VQLA~\cite{bai2023surgical}              &$\mathcal{L}_1$+GIoU&CE&   $\times$          & 0.6206          & 0.2742          & 0.7731          & 0.4562         & 0.3317          & 0.7652          \\
Surgical-VQLA++                   &GIoU&Focal& \textbf{\checkmark}        & \textbf{0.6568} & \textbf{0.3180} & \textbf{0.7982} & \textbf{0.5071} & \textbf{0.4476} & \textbf{0.7758} \\
\bottomrule[1pt] 
\end{tabular}}
\end{table*}

\begin{table*}[t]
\caption{Comparison experiments of our Surgical-VQLA++ against SOTA solutions. We implement all the model on the extended datasets, so the reported results will be different from the results reported in their original publications.
GF: Gated Fusion, CA: Co-Attention layer, MCC: Multimodal Collaborated Calibration, GCC: Global Contextual Calibration, C$^2$G: C$^2$G-ViL Embedding.}
\centering
\label{tab:main}  
\resizebox{0.85\textwidth}{!}{
\begin{tabular}{c|ccc|ccc}
\toprule[1pt]
\multirow{2}{*}{Models}            & \multicolumn{3}{c|}{EndoVis-18-VQLA}                       & \multicolumn{3}{c}{EndoVis-17-VQLA}                       \\ \cline{2-7}
& Acc & F-Score & mIoU & Acc & F-Score & mIoU              \\ \hline
VisualBERT~\cite{li2019visualbert}        & 0.6203          & 0.3061          & 0.7433          & 0.4647          & 0.2471          & 0.7336          \\
VisualBERT ResMLP~\cite{seenivasan2022surgical} & 0.6433          & 0.3218          & 0.7576          & 0.4633          & 0.2674          & 0.7387          \\
MCAN~\cite{yu2019deep}              & 0.6366          & 0.3191          & 0.7827          & 0.4506          & 0.2640          & 0.7647          \\
VQLA-DeiT~\cite{touvron2021training}         & 0.6183          & 0.2827          & 0.7343          & 0.4379          & 0.2324          & 0.7224          \\
MUTAN~\cite{ben2017mutan}             & 0.6343          & 0.2930          & 0.7764          & 0.4817          & 0.2900          & 0.7690          \\
MFH~\cite{yu2018beyond}               & 0.6241          & 0.3149          & 0.7679          & 0.4774         & 0.3418          & 0.7499          \\
BlockTucker~\cite{ben2019block}       & 0.6391          & 0.3077          & 0.7693          & 0.4520          & 0.3059          & 0.7640          \\
CAT-ViL DeiT~\cite{bai2023co}      & 0.6290          & 0.2807          & 0.7776          & 0.4887          & 0.2812          & 0.7657          \\
Surgical-VQLA~\cite{bai2023surgical}         & 0.6456          & \textbf{0.3219} & 0.7212          & 0.4322          & 0.2416          & 0.7321          \\ \hline
Surgical-VQLA++ w/o C$^2$G &  0.6249 & 0.3069 & 0.7578 & 0.4548 & 0.3325 & 0.7441 \\
Surgical-VQLA++ w/o CA                                                                         & 0.6398          & 0.2998         & 0.7808          & 0.4407          & 0.3887          & 0.7569          \\
Surgical-VQLA++ w/o GF                                                                       & 0.6349           & 0.2814         & 0.7742          & 0.4581          & 0.3464          & 0.7677          \\
Surgical-VQLA++ w/o MCC & 0.6362&	0.3187&	0.7740&	0.4909&	0.3401&	0.7675 \\
Surgical-VQLA++ w/o GCC &0.6314 &0.3210 &0.7845 & 0.4477 & 0.3282 & 0.7655 \\
\hline
Surgical-VQLA++      & \textbf{0.6568} & 0.3180          & \textbf{0.7982} & \textbf{0.5071} & \textbf{0.4476} & \textbf{0.7758} \\
\bottomrule[1pt] 
\end{tabular}}
\end{table*}

\subsection{Implementation Details}
Our proposed solution is compared against the following models: VisualBERT\footnote{\href{https://github.com/uclanlp/visualbert}{github.com/uclanlp/visualbert}}~\cite{li2019visualbert}, VisualBERT ResMLP\footnote{\href{https://github.com/lalithjets/surgical_vqa}{github.com/lalithjets/surgical\_vqa}}~\cite{seenivasan2022surgical}, VQLA-DeiT\footnote{\href{https://github.com/rwightman/pytorch-image-models}{github.com/rwightman/pytorch-image-models}}~\cite{touvron2021training}, MCAN\footnote{\href{https://github.com/MILVLG/mcan-vqa}{github.com/MILVLG/mcan-vqa}}~\cite{yu2019deep}, MUTAN\footnote{\href{https://github.com/Cadene/block.bootstrap.pytorch}{github.com/Cadene/block.bootstrap.pytorch}}~\cite{ben2017mutan}, MFH\footnote{\href{https://github.com/MILVLG/openvqa}{github.com/MILVLG/openvqa}}~\cite{yu2018beyond}, BlockTucker\footnotemark[6]~\cite{ben2019block}, CAT-ViL DeiT\footnote{\href{https://github.com/longbai1006/CAT-ViL}{github.com/longbai1006/CAT-ViL}}~\cite{bai2023co}, and our earlier version Surgical-VQLA\footnote{\href{https://github.com/longbai1006/Surgical-VQLA}{github.com/longbai1006/Surgical-VQLA}}~\cite{bai2023surgical}. We employ the same loss $\mathcal{L}_{VQLA}$ for all models to keep a fair comparison. Following their official repositories, we implement all the model on the extended datasets, so the reported results will be different from the results reported in their original publications.

All models are trained using the Python PyTorch framework on NVIDIA RTX 3090 GPU. Adam optimizer~\cite{kingma2014adam} is employed for $80$ epochs with a batch size of $64$ and a learning rate of $1 \times 10^{-5}$.

\begin{table*}[t]
\caption{Comparison of various test samples on Surgical-VQLA, CAT-ViL DeiT, and our Surgical-VQLA++. The four types of samples include the tissue being operated on, the current state of the instruments, the instrument location in the operating area, and the instrument identification. Please note that there are no QA pairs related to tissues in the EndoVis-17-VQLA dataset.}
\centering
\label{tab:break} 
\resizebox{0.95\textwidth}{!}{
\begin{tabular}{c|c|ccc|ccc}
\hline
                        &                             & \multicolumn{3}{c|}{EndoVis-18-VQLA}                                            & \multicolumn{3}{c}{EndoVis-17-VQLA}                                                                     \\ \cline{3-8} 
\multirow{-2}{*}{Model} & \multirow{-2}{*}{Data Type} & Acc             & F-Score         & mIoU                   & Acc             & F-Score                                 & mIoU                   \\ \hline
CAT-ViL DeiT~\cite{bai2023co}            &                             & 1.0000          & 1.0000       & 0.6910±0.1558          & /               & /                                          & /                      \\
Surgical-VQLA~\cite{bai2023surgical}            &                             & 1.0000          & 1.0000      & 0.7253±0.1178 & /               & /                                           & /                      \\
Surgical-VQLA++         & \multirow{-3}{*}{Tissue}     & 1.0000          & 1.0000      & \textbf{0.8124±0.1234}          & /               & /                                         & /                      \\ \hline
CAT-ViL DeiT~\cite{bai2023co}            &                             & 0.5426          & 0.1646       & 0.8279±0.0919          & \textbf{0.3305} & 0.1898                              & 0.7656±0.0988          \\
Surgical-VQLA~\cite{bai2023surgical}            &                             & 0.5049          & 0.1676        & 0.8174±0.0999          & 0.3051          & 0.2882                               & 0.7479±0.0701          \\
Surgical-VQLA++         & \multirow{-3}{*}{Action}     & \textbf{0.5771} & \textbf{0.2022}  & \textbf{0.8655±0.0859} & 0.2966          & \textbf{0.3336}      & \textbf{0.7972±0.0967} \\ \hline
CAT-ViL DeiT~\cite{bai2023co}            &                             & 0.5585          & 0.4802          & 0.7491±0.2122          & 0.5297          & 0.4988                                & 0.7182±0.2521          \\
Surgical-VQLA~\cite{bai2023surgical}           &                             & \textbf{0.6117} & \textbf{0.5616} & 0.6411±0.1908          & 0.5424 & 0.5269                       & 0.6450±0.1787          \\
Surgical-VQLA++         & \multirow{-3}{*}{Location}     & 0.6064          & 0.4678         & \textbf{0.7515±0.2145} & \textbf{0.6441}          & \textbf{0.6181}              & \textbf{0.7187±0.2535} \\ \hline
CAT-ViL DeiT~\cite{bai2023co}            &                             & \textbf{0.6791}          & \textbf{0.2266} & \textbf{0.7911±0.1421}          & 0.5763          & \textbf{0.4664}           & \textbf{0.8176±0.1369}          \\
Surgical-VQLA~\cite{bai2023surgical}            &                             & 0.6722 & 0.2067        & 0.7039±0.1358          & 0.5593          & 0.4002      & 0.7166±0.1290          \\
Surgical-VQLA++         & \multirow{-3}{*}{Instrument}     & 0.6543          & 0.2194   &    0.7820±0.1441 & \textbf{0.5805} & 0.4173  & 0.8116±0.1399 \\ \hline
\end{tabular}}
\end{table*}

\subsection{Results}
We first compare our proposed method with our preliminary version, Surgical-VQLA~\cite{bai2023surgical}. The results of this comparison are presented in Table~\ref{tab:ab1}. To ensure a fair evaluation, we re-implement the Surgical-VQLA model using our extended dataset. Through the introduction of our Surgical-VQLA++ framework, we observe improvements in the metrics for both datasets, ranging from 1.06\% to 11.59\%. These enhancements affirm the significant improvement our work has made in enhancing the VQLA performance of the model. 

Next, we perform a quantitative evaluation against SOTA methods, as shown in Table~\ref{tab:main}. In order to maintain fairness, all models utilize uncertainty loss, employing focal loss for QA prediction and GIoU for the localization task. Our proposed methodology demonstrates exceptional performance across accuracy and mIoU metrics, surpassing all baseline models. This substantiates the efficacy and superiority of our approach. Notably, the performance of Surgical-VQLA exhibits improvement compared to the results presented in Table~\ref{tab:ab1}, which suggests the broad applicability of the adopted loss function strategy.

Moreover, we assess the effects of removing the (i) co-attention layers, (ii) the gated fusion module, (iii) multimodal collaborated calibration, (iv) global contextual calibration, and (v) C$^2$G-ViL embedding, respectively. Experimental results in Table~\ref{tab:main} conducted on the EndoVis-18-VQLA dataset reveal that the absence of any of these modules leads to a significant degradation in model performance. Similarly, in the external validation set, EndoVis-17-VQLA, both accuracy and mIoU decrease significantly. These observations indicate that all three proposed modules positively contribute to achieving the best results.

Furthermore, we provide a detailed evaluation of the model's performance across different types of test samples, as shown in Table~\ref{tab:break}. These samples are divided based on the types of questions, including tissue recognition, action state, instrument location, and instrument identification. Firstly, all models achieve full scores in tissue recognition as this task setup is not complicated. Secondly, our model achieves the best performance on action state recognition and instrument location questions, while it performs unsatisfactorily in instrument identification. However, although the model does not perform the best in instrument-related questions during text dialogue, its localization module maintains a high level of bounding box detection performance. These findings underscore the effectiveness of our framework in handling diverse surgical scenarios and questions, thus enhancing its applicability and reliability in real-world settings.

\begin{table*}[t]
\caption{Ablation study on different visual feature extractors.}
\centering
\label{tab:visual feature}  
\resizebox{0.85\textwidth}{!}{
\begin{tabular}{c|ccc|ccc|ccc}
\toprule[1pt]
Model &
\multicolumn{3}{c|}{Visual Feature Extractor}   & \multicolumn{3}{c|}{EndoVis-18-VQLA}     & \multicolumn{3}{c}{EndoVis-17-VQLA}                                                            \\ \hline
\multirow{3}{*}{Ours} & Detection & Feature Extraction                      & Inference Speed   & Acc & \multicolumn{1}{c}{F-Score} & \multicolumn{1}{c|}{mIoU} & Acc & \multicolumn{1}{c}{F-Score} & \multicolumn{1}{c}{mIoU} \\ \cline{2-10}
&$\times$ & ResNet~\cite{he2016deep} & 150.6 FPS & \multicolumn{1}{r}{\textbf{0.6568}} & 0.3180 & \textbf{0.7982} & \multicolumn{1}{r}{\textbf{0.5071}} & \textbf{0.4476}  & \textbf{0.7758}  \\
&FRCNN~\cite{ren2015faster}  &ResNet~\cite{he2016deep}  & 18.09 FPS & \multicolumn{1}{r}{0.6340}  & \textbf{0.3720} & 0.7218 & \multicolumn{1}{r}{0.4251} & 0.2946 & 0.7050 \\       
\bottomrule[1pt] 
\end{tabular}}
\end{table*}

\begin{table*}[t]
\caption{Ablation study on different loss function combinations. For QA prediction, we use the CE loss and focal loss~\cite{lin2017focal}. For the localization task, we explore the effects of GIoU~\cite{rezatofighi2019generalized} and $\mathcal{L}_1 +$ GIoU~\cite{carion2020end}. The integration of uncertainty loss~\cite{kendall2018multi} is also introduced.}
\centering
\label{tab:loss combinations1}  
\resizebox{0.85\textwidth}{!}{
\begin{tabular}{ccc|ccc|ccc}
\toprule[1pt]
\multirow{2}{*}{Localization} & \multirow{2}{*}{Classification} & \multirow{2}{*}{Uncertainty} & \multicolumn{3}{c|}{EndoVis-18-VQLA}                       & \multicolumn{3}{c}{EndoVis-17-VQLA}                       \\ \cline{4-9}
                                &                                 &                              & Acc               & F-Score           & mIoU              & Acc               & F-Score           & mIoU              \\ \hline
GIoU                            & CE                              & $\times$                          & 0.6199          & 0.3048          & 0.7613          & 0.4929          & 0.2839          & 0.7519          \\
$\mathcal{L}_1$+GIoU                         & CE                              & $\times$                           & 0.6295          & 0.2939          & 0.7737           & 0.5113          & 0.2976          & 0.7570          \\
GIoU                            & Focal                           & $\times$                           & 0.6433          & \textbf{0.3793} & 0.7885          & 0.4379          & 0.4396 & 0.7709          \\
$\mathcal{L}_1$+GIoU                         & Focal                           & $\times$                            & 0.6250          & 0.3610          & 0.7725          & 0.5042          & 0.3865          & 0.7394          \\
GIoU                            & CE                              & \checkmark                            & 0.6267          & 0.2917          & 0.7620          & \textbf{0.5170} & 0.3709          & 0.7625          \\
$\mathcal{L}_1$+GIoU                         & CE                              & \checkmark                          & 0.6211           & 0.2947          & 0.7795          & 0.5127          & 0.3563          & 0.7691          \\
\textbf{GIoU}                            & \textbf{Focal}                           & \textbf{\checkmark}                            & \textbf{0.6568} & 0.3180          & \textbf{0.7982} & 0.5071          & \textbf{0.4476}          & \textbf{0.7758} \\
$\mathcal{L}_1$+GIoU                         & Focal                           & \checkmark                           & 0.6371           & 0.2998          & 0.7712          & 0.5071          & 0.3078          & 0.7530          \\ 
\bottomrule[1pt]   
\end{tabular}}
\end{table*}

\begin{table*}[]
\caption{Ablation study on different IoU loss functions.}
\centering
\label{tab:loss combinations2}  
\resizebox{0.85\textwidth}{!}{
\begin{tabular}{ccc|ccc|ccc}
\toprule[1pt]
\multirow{2}{*}{Localization} & \multirow{2}{*}{Classification} & \multirow{2}{*}{Uncertainty} & \multicolumn{3}{c|}{EndoVis-18-VQLA}                       & \multicolumn{3}{c}{EndoVis-17-VQLA}                       \\ \cline{4-9}
                                &                                 &                              & Acc               & F-Score           & mIoU              & Acc               & F-Score           & mIoU              \\ \hline
IoU  & Focal & \textbf{\checkmark} & 0.6122          & \textbf{0.3634} & 0.4097          & 0.4361           & 0.4408          & 0.4104          \\
CIoU~\cite{zheng2020distance} & Focal & \textbf{\checkmark} & 0.6255          & 0.3236          & 0.7859          & 0.4590           & \textbf{0.4486} & 0.7640          \\
DIoU~\cite{zheng2020distance} & Focal & \textbf{\checkmark} & 0.6333          & 0.3271          & 0.7561          & 0.4618          & 0.2572          & 0.7552          \\
\textbf{GIoU}~\cite{rezatofighi2019generalized} & \textbf{Focal} & \textbf{\checkmark} & \textbf{0.6568} & 0.3180          & \textbf{0.7982} & \textbf{0.5071} & 0.4476          & \textbf{0.7758} \\ 
\bottomrule[1pt]   
\end{tabular}}
\end{table*}

\begin{table}[t]
\caption{Ablation study on different fusion strategies. All experiments employ the same feature extractor, DeiT backbone, and prediction heads. "Attn": Attention.}
\centering
\label{tab:fusion strategies}  
\resizebox{0.48\textwidth}{!}{
\begin{tabular}{c|ccc|ccc}
\toprule[1pt]
\multirow{2}{*}{Models} & \multicolumn{3}{c|}{EndoVis-18-VQLA}                       & \multicolumn{3}{c}{EndoVis-17-VQLA}                       \\    \cline{2-7}
                        & Acc               & F-Score           & mIoU              & Acc               & F-Score           & mIoU              \\ \hline
Concatenation~\cite{li2019visualbert}           & 0.6183          & 0.2827          & 0.7343          & 0.4379          & 0.2324          & 0.7224          \\
JCA~\cite{praveen2022joint}                     & 0.6221          & 0.2895          & 0.7542          & 0.4463          & 0.2558          & 0.7465          \\
MMHCA~\cite{georgescu2023multimodal}                   & 0.6295          & 0.2798          & 0.7700          & 0.4364          & 0.3006          & 0.7579          \\
MAT~\cite{wu2022multimodal}                     & 0.6303          & 0.2872          & 0.7632          & 0.4393          & 0.2942          & 0.7547          \\
GMU~\cite{arevalo2017gmu}            & 0.6361          & 0.3490          & 0.7196          & 0.4167          & 0.2360          & 0.7326          \\
Self-Attn~\cite{vaswani2017attention}               & 0.6262          & 0.2810          & 0.7714          & 0.4449          & 0.3030          & 0.7548          \\
Guided-Attn~\cite{yu2019deep}            & 0.6295          & 0.2983          & 0.7766          & 0.4633          & 0.3018          & 0.7582          \\
Co-Attn (Bi)            & 0.6262          & 0.2915          & 0.7661          & 0.4463          & 0.3033          & 0.7482          \\
Co-Attn (V2T)           & 0.6308          & 0.3037          & 0.7510          & 0.4040          & 0.2201          & 0.7341          \\
Co-Attn (T2V)           & 0.6366          & 0.2858          & 0.7791          & 0.4407          & 0.2569          & 0.7563          \\
Ours with Self-Attn    & 0.6484          & 0.3447          & 0.7749          & 0.4576          & 0.2545          & 0.7543          \\
Ours with Guided-Attn  & 0.6403          & 0.2817          & 0.7886          & 0.4845          & 0.3183          & 0.7609          \\
Ours with Co-Attn (Bi)       & 0.6251          & 0.3110          & 0.7679          & 0.4393          & 0.2768          & 0.7498          \\
Ours with Co-Attn (V2T)           & 0.6233          & \textbf{0.3602} & 0.7609          & 0.4703          & \textbf{0.4666} & 0.7517          \\
Ours with Co-Attn (T2V)           & \textbf{0.6568} & 0.3180          & \textbf{0.7982} & \textbf{0.5071} & 0.4476          & \textbf{0.7758}                     \\
\bottomrule[1pt]
\end{tabular}}
\end{table}

\begin{table}[t]
\caption{Ablation study on the number of co-attention layers.}
\centering
\label{tab:co-attention layers}  
\resizebox{0.48\textwidth}{!}{
\begin{tabular}{c|ccc|ccc}
\toprule[1pt]
\multirow{2}{*}{Co-attention Layer} & \multicolumn{3}{c|}{EndoVis-18-VQLA}                       & \multicolumn{3}{c}{EndoVis-17-VQLA}                       \\   \cline{2-7}
                                    & Acc               & F-Score           & mIoU              & Acc               & F-Score           & mIoU              \\ \hline
2                                   & 0.6211           & 0.2754          & 0.7829            & 0.4520          & 0.3285          & 0.7706          \\
4                                   & 0.6260          & 0.3301          & 0.7829          & 0.4379          & 0.3554          & 0.7685          \\
\textbf{6}                                   & \textbf{0.6568} & 0.3180          & \textbf{0.7982} & \textbf{0.5071} & 0.4476          & \textbf{0.7758} \\
8                                   & 0.6252          & \textbf{0.3683} & 0.7772          & 0.4025          & \textbf{0.4555} & 0.7664          \\
10                                  & 0.5641          & 0.2126          & 0.7049          & 0.4633          & 0.2487          & 0.7167     \\    
\bottomrule[1pt]   
\end{tabular}}
\end{table}

\begin{table}[t]
\caption{Ablation on the weight parameters of the adversarial perturbations. $\alpha$ and $\beta$ denote the perturbation weight on text and visual embedding. The bolded weights are the parameters we use in all the above experiments. "w/o CTAS" denotes our model without contrastive training using adversarial samples.}
\centering
\label{tab:contrastive training}  
\resizebox{0.48\textwidth}{!}{
\begin{tabular}{cc|ccc|ccc}
\toprule[1pt]
\multirow{2}{*}{$\alpha$} & \multirow{2}{*}{$\beta$} & \multicolumn{3}{c|}{EndoVis-18-VQLA}                       & \multicolumn{3}{c}{EndoVis-17-VQLA}                       \\ \cline{3-8}
                         &                            & Acc               & F-Score           & mIoU              & Acc               & F-Score           & mIoU              \\ \hline
\multicolumn{2}{c|}{w/o CTAS}                         & 0.6419          & 0.3378           & 0.7787          & 0.4689          & 0.3658           & 0.7705          \\
0.5                      & 1                          & 0.6398          & 0.3243          & 0.7872          & 0.4562          & 0.4366          & 0.7716          \\
0.5                      & 0.5                        & 0.6201          & 0.3118          & 0.7881          & 0.4548          & 0.4185          & 0.7675          \\
0.5                      & 0.1                        & 0.6416          & 0.3376          & 0.7892          & 0.4802          & 0.3043          & 0.7592          \\
1                       & 1                          & 0.6359          & 0.2938          & 0.7675          & 0.4884          & 0.4373 & 0.7599          \\
\textbf{1}                        & \textbf{0.5}                        & \textbf{0.6568} & 0.3180          & \textbf{0.7982} & \textbf{0.5071} & \textbf{0.4476} & \textbf{0.7758} \\
1                        & 0.1                        & 0.6529           & 0.3066          & 0.7838          & 0.4703          & 0.3994          & 0.7621          \\
2                        & 1                          & 0.6584          & 0.3566          & 0.7786          & 0.4590          & 0.4025          & 0.7477          \\
2                        & 0.5                        & 0.6232          & \textbf{0.3599} & 0.7829          & 0.4619          & 0.3999          & 0.7711          \\
2                        & 0.1                        & 0.6317          & 0.3560          & 0.7827          & 0.4958          & 0.3960          & 0.7682  \\    
\bottomrule[1pt]   
\end{tabular}}
\end{table}

\subsection{Ablation Study}
\noindent \textbf{Effects of the visual feature extractors.} 
In Table~\ref{tab:visual feature}, a comparative analysis is presented, evaluating the model's performance before and after the incorporation of object proposals using global feature extractor ResNet18~\cite{he2016deep}. Apart from the significant improvement in inference speed brought by the end-to-end architecture, our model also demonstrates superior quantitative performance. This can be attributed to the global feature extractor's ability to avoid directing attention to incorrect regions as a result of erroneous detections.

\noindent \textbf{Effects of loss function combinations.}
In Table~\ref{tab:loss combinations1}, we present the analysis of various combinations of loss functions and their impact on the performance of our proposed model. When compared to using GIoU alone, the combination of $\mathcal{L}_1$ and GIoU does not consistently improve the evaluation metrics, indicating that it is not a superior strategy. Conversely, the integration of focal loss exhibits notable superiority over CE Loss. Additionally, the incorporation of uncertainty loss~\cite{kendall2018multi} further enhances the model's performance, especially when combined with GIoU and focal loss. Furthermore, we investigate several advanced IoU functions in Table~\ref{tab:loss combinations2} using the aforementioned combination strategies. The results reveal that, compared with DIoU and CIoU~\cite{zheng2020distance}, GIoU loss is the most suitable choice for our multitask learning framework.

\begin{figure*}[]
    \centering\includegraphics[width=0.88\linewidth, trim=0 10 0 0]{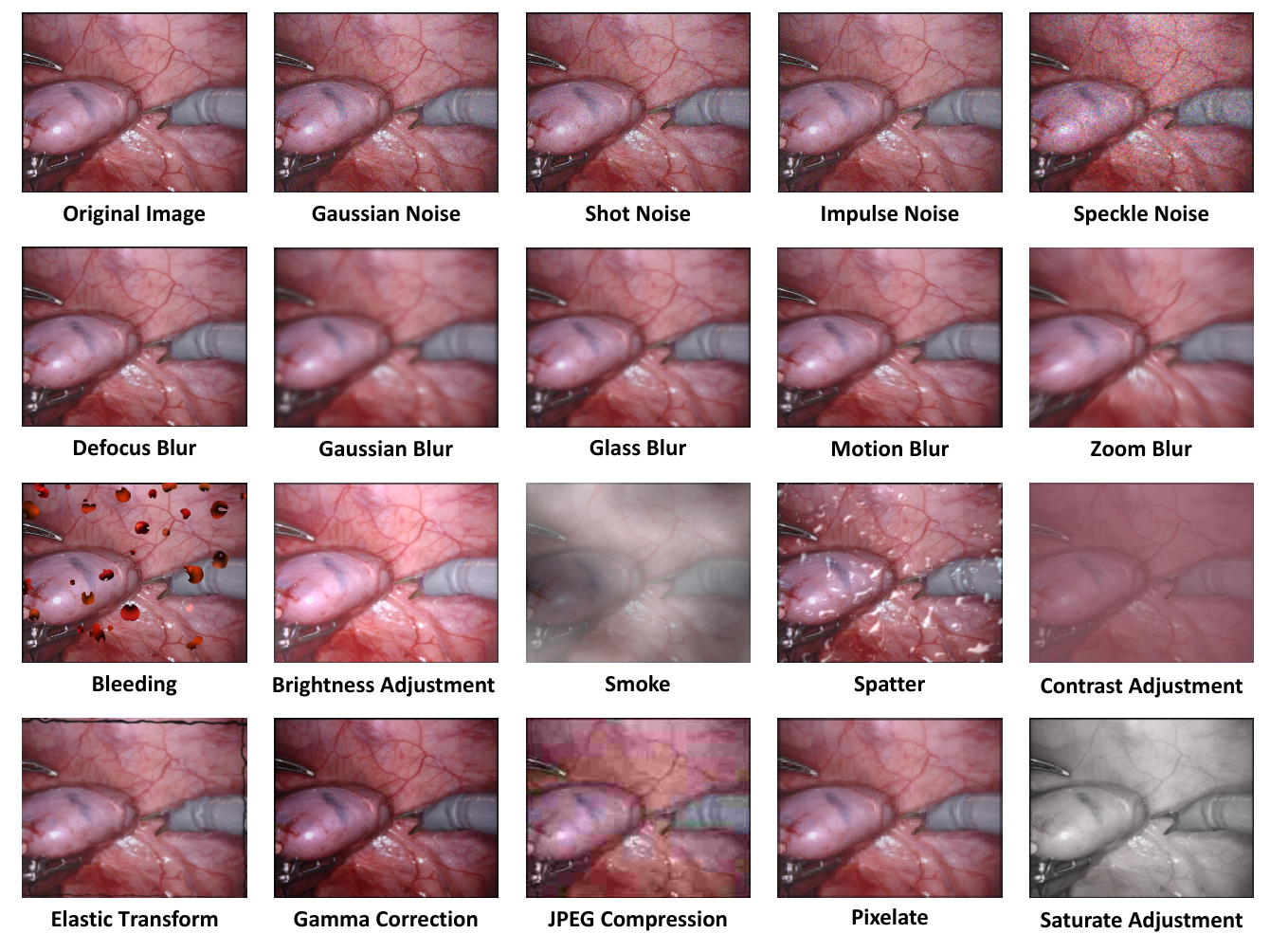}
    \caption{Corrupted data visualization for the robustness assessment. We simulate four different types of image corruption, including noise, blur, occlusion, and digital damage.}
    \label{fig:supply}
\end{figure*}

\noindent \textbf{Effects of different fusion strategies.}
In Table~\ref{tab:fusion strategies}, we investigate the impact of different ViL embedding while keeping the DeiT backbone and feature extractors consistent. The comparisons include Concatenation~\cite{li2019visualbert}, JCA~\cite{praveen2022joint}, MMHCA~\cite{georgescu2023multimodal}, MAT~\cite{wu2022multimodal}, GMU~\cite{arevalo2017gmu}, Self-Attention~\cite{vaswani2017attention}, Guided-Attention~\cite{yu2019deep}, and Co-Attention (T2V: Use text to guide vision~\cite{yu2019deep}; V2T: Use vision to guide text; Bi: Bidirectional guidance). Additionally, we explore our Surgical-VQLA++ solution using the attention mechanisms mentioned above for a comprehensive comparison. The study demonstrates that our ViL embedding strategy outperforms other sophisticated methods. Moreover, in comparison to other forms of attention, utilizing text information to guide visual feature embedding significantly helps the model focus on crucial regions within an image, resulting in outstanding performance in VQLA tasks.

\noindent \textbf{Effects of the number of co-attention layers.} In Table~\ref{tab:co-attention layers}, we evaluate the impact of different co-attention configurations on our model's performance. After careful analysis, we identify layer number $=6$ as the most suitable co-attention configuration for our final model. Deviations from this optimal value result in a slight decrease in performance, with a significant degradation observed when the layer number exceeds 8.

\noindent \textbf{Effects of the weights in adversarial perturbation.}
In Table~\ref{tab:contrastive training}, we systematically investigate the influence of weight parameters $\alpha$ and $\beta$ in contrastive training on the overall performance of our model. Through a grid search, we determine that setting $\alpha=1$ and $\beta = 0.5$ yields the best performance. These weight parameters are crucial for achieving optimal results in our proposed framework. Besides, as illustrated in the first row of Table~\ref{tab:contrastive training}, we report the results obtained when the contrastive training strategy is excluded. The performance exhibits a significant degradation, providing further evidence to support the effectiveness of our proposed training strategy.

\subsection{Robustness against Image Corruption}
We conducted a comprehensive robustness assessment to evaluate how effectively our model handles challenging image corruption. We used a benchmark consisting of 19 different types of image corruption, with 18 sourced from a widely recognized benchmark~\cite{hendrycks2019benchmarking} and 1 simulating blood occlusion on the camera lens during surgery~\cite{garcia2021image}. The visualization of these nineteen image corruption techniques is shown in Figure~\ref{fig:supply}. All corruption methods are categorized into five severity levels to observe the model's performance under varying degrees of corruption. Below is a brief introduction to these four categories and 19 types of image corruption.

The `Noise' category represents various types of noise that may occur during image capture, storage, and transmission. For instance, low-light conditions typically cause Gaussian noise. Impulse noise, a color variation of salt-and-pepper noise, can be generated by bit errors. Shot noise is a type of electronic noise. Speckle noise is an additive noise that varies based on the intensity of the original pixel.

The `Blur' category represents image blur that can occur due to defocusing or motion during capture and transmission. Defocus blur appears when the image is out of focus. Gaussian Blur occurs when surgical images are excessively de-noised. Glass Blur usually results from optical distortions or errors in compression/ transmission. Motion blur occurs when the camera moves very fast. Zoom blur appears when a camera rapidly moves toward an object.

The `Occlusion' category represents reductions in image quality due to various objective conditions in real surgical scenes, such as abnormal lighting and obstructions. Bleeding mimics blood occlusion on the lens in surgical scenes. Brightness varies with surgical light intensity. Smoke, commonly seen during cauterization in surgery, shrouds the object. Spatter simulates water occluding the camera.

The `Digital' category represents digital damage that can occur during image collection and transmission due to hardware issues. Contrast is influenced by the color of the tissues and lighting conditions. Elastic transform can stretch or contract the surgical area. Gamma Correction adjusts the brightness and contrast of the surgical scene through a nonlinear transformation. JPEG is a compression artifact effect produced by the lossy compression of surgical images. Pixelation occurs when upsampling low-resolution surgical images. Saturate Adjustment changes the purity of image colors.

\begin{figure*}[b]
    \centering
    \includegraphics[width=\linewidth]{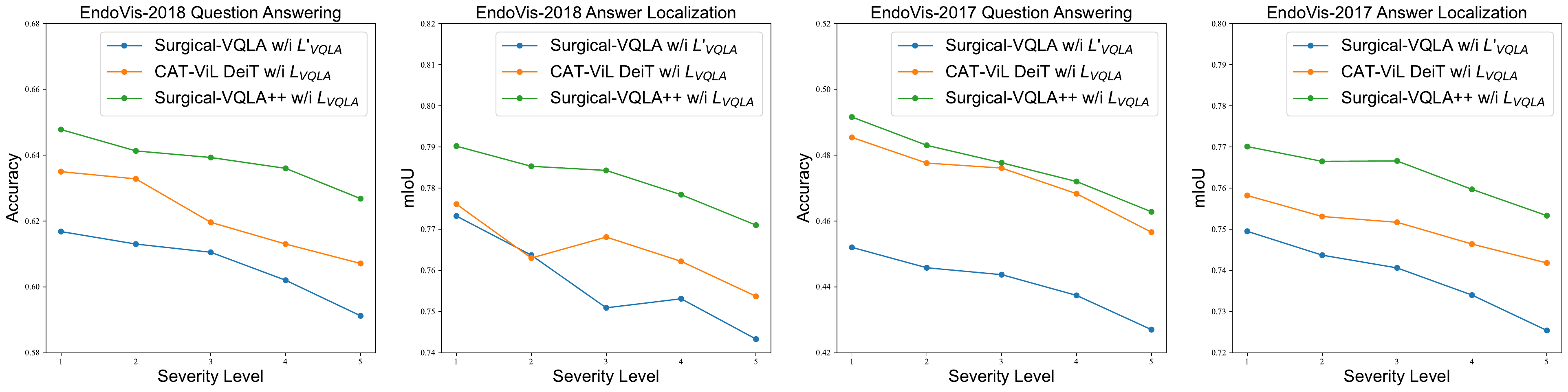}
    \caption{Robustness comparison on each severity level. We process the data using 19 image corruption approaches with 5 severity levels, and average the predictions on each severity level.}
    \label{fig:robust}
\end{figure*}

Then, we directly conduct inference on the corrupted test dataset and calculate the average results across the different corruption techniques for each severity level. The outcomes of our comprehensive robustness experiments, categorized by the severity level, are depicted in Figure~\ref{fig:robust}. A consistent deterioration in performance is observed as the severity level of corruption escalates. Furthermore, Table~\ref{tab:robustness} provides an overview of the average results for each corruption type. Our Surgical-VQLA++ showcases notable performance advantages across various corruption scenarios. It is worth highlighting that our methods consistently achieve higher accuracy and mIoU compared to both the original Surgical-VQLA and CAT-ViL approaches.
While our method and CAT-ViL DeiT exhibit comparable performance in the QA task of EndoVis-17-VQLA, our approach still maintains an overall advantage. In this context, the incorporation of feature calibration and adversarial contrastive training effectively enhances the robustness of our model in dealing with image corruption challenges.

\begin{table*}[t]
\caption{Robustness comparison on Surgical-VQLA, CAT-ViL DeiT, and our Surgical-VQLA++. We compute the average results on each corruption type.}
\centering
\label{tab:robustness}  
\resizebox{1.0\textwidth}{!}{
\begin{tabular}{c|c|ccc|ccc|ccc|ccc}
\toprule[1pt]
\multicolumn{2}{c|}{Dataset}             & \multicolumn{6}{c|}{EndoVis-18-VQLA}                                                                           & \multicolumn{6}{c}{EndoVis-17-VQLA}                                                                           \\ \hline
\multicolumn{2}{c|}{Metric}              & \multicolumn{3}{c|}{Acc}                               & \multicolumn{3}{c|}{mIoU}                              & \multicolumn{3}{c|}{Acc}                               & \multicolumn{3}{c}{mIoU}                              \\ \hline
\multicolumn{2}{c|}{Model}               & \multicolumn{1}{c}{\makecell[c]{Surgical- \\ VQLA~\cite{bai2023surgical}}} & \multicolumn{1}{c}{\makecell[c]{CAT-ViL \\ DeiT~\cite{bai2023co}}}   & \multicolumn{1}{c|}{\makecell[c]{Surgical- \\ VQLA++}}   & \multicolumn{1}{c}{\makecell[c]{Surgical- \\ VQLA~\cite{bai2023surgical}}}  & \multicolumn{1}{c}{\makecell[c]{CAT-ViL \\ DeiT~\cite{bai2023co}}} & \multicolumn{1}{c|}{\makecell[c]{Surgical- \\ VQLA++}}  & \multicolumn{1}{c}{\makecell[c]{Surgical- \\ VQLA~\cite{bai2023surgical}}}  & \multicolumn{1}{c}{\makecell[c]{CAT-ViL \\ DeiT~\cite{bai2023co}}}  & \multicolumn{1}{c|}{\makecell[c]{Surgical- \\ VQLA++}}  & \multicolumn{1}{c}{\makecell[c]{Surgical- \\ VQLA~\cite{bai2023surgical}}}  & \multicolumn{1}{c}{\makecell[c]{CAT-ViL \\ DeiT~\cite{bai2023co}}}  & \multicolumn{1}{c}{\makecell[c]{Surgical- \\ VQLA++}} \\ \hline
\multicolumn{2}{c|}{Loss}               & $\mathcal{L}'_{VQLA}$  & $\mathcal{L}_{VQLA}$   & $\mathcal{L}_{VQLA}$   & $\mathcal{L}'_{VQLA}$   & $\mathcal{L}_{VQLA}$   & $\mathcal{L}_{VQLA}$  & $\mathcal{L}'_{VQLA}$ & $\mathcal{L}_{VQLA}$   & $\mathcal{L}_{VQLA}$  & $\mathcal{L}'_{VQLA}$   & $\mathcal{L}_{VQLA}$  & $\mathcal{L}_{VQLA}$  \\ \hline
\multirow{4}{*}{Noise}     & Gauss.     & 0.5927            & 0.6180      & \textbf{0.6396 }             & 0.7657            & 0.7753      & \textbf{0.7929}              & 0.4435            & 0.4663      & \textbf{0.4814}              & 0.7581            & 0.7638      & \textbf{0.7775}              \\ \cline{2-14}
& Shot       & 0.5905            & 0.6213      & \textbf{0.6464}              & 0.7657            & 0.7760      & \textbf{0.7949}              & 0.4386            & 0.4711      & \textbf{0.4740}              & 0.7605            & 0.7640      & \textbf{0.7790 }             \\\cline{2-14}
& Impulse    & 0.6023            & 0.6233      & \textbf{0.6407}              & 0.7672            & 0.7768      & \textbf{0.7951}              & 0.4369            & 0.4712      & \textbf{0.4749}              & 0.7576            & 0.7638      & \textbf{0.7802}              \\\cline{2-14}
& Speckle    & 0.5719            & 0.5915      & \textbf{0.6096}              & 0.7398            & 0.7427      & \textbf{0.7599}              & 0.4404            & \textbf{0.4756}      & 0.4339              & 0.7300            & 0.7390      & \textbf{0.7508}              \\ \hline
\multirow{5}{*}{Blur}      & Defocus    & 0.6188            & 0.6299      & \textbf{0.6523}              & 0.7760            & 0.7787      & \textbf{0.7932}              & 0.4433            & 0.4686      & \textbf{0.4938}              & 0.7515            & 0.7588      & \textbf{0.7711}              \\ \cline{2-14}
                           & Gauss.     & 0.6006            & 0.6157      & \textbf{0.6474}              & 0.7483            & 0.7471      & \textbf{0.7678}              & 0.4485            & \textbf{0.4720}      & 0.4647              & 0.7090            & 0.7281      & \textbf{0.7446}              \\\cline{2-14}
                           & Glass      & 0.6236            & \textbf{0.6584}      & 0.6561              & 0.7763            & 0.7804      & \textbf{0.7938}              & 0.4565            & 0.4799      & \textbf{0.4949}              & 0.7607            & 0.7632      & \textbf{0.7736}              \\\cline{2-14}
                           & Motion     & 0.6135            & 0.6314      & \textbf{0.6567}              & 0.7730            & 0.7771      & \textbf{0.7904}              & 0.4480            & 0.4743      & \textbf{0.4972}              & 0.7527            & 0.7610      & \textbf{0.7729}              \\\cline{2-14}
                           & Zoom       & 0.5880            & 0.6249      & \textbf{0.6292}              & 0.7293            & \textbf{0.7690}      & 0.7683              & 0.4362            & \textbf{0.4734}      & 0.4556              & 0.7330            & 0.7485      & \textbf{0.7557}              \\ \hline
\multirow{4}{*}{Occlusion} & Bleeding &0.5944 &0.5978 &\textbf{0.6080} 
                            & 0.7361 & 0.7600 & \textbf{0.7736}
                            & 0.4391 & 0.4468 & \textbf{0.4576}
                            & 0.7083 & 0.7336 & \textbf{0.7573}
                            \\ \cline{2-14}
                           & Brightness & 0.6275            & 0.6242      & \textbf{0.6444}              & 0.7656            & 0.7725      & \textbf{0.7921}              & 0.4417            & 0.4708      & \textbf{0.4975}              & 0.7488            & 0.7566      & \textbf{0.7728}              \\ \cline{2-14}
                           & Smoke        & \textbf{0.6333}            & 0.6213      & 0.6318              & 0.7597            & 0.7606      & \textbf{0.7687}              & 0.4302            & \textbf{0.4729}      & 0.4636              & 0.7239            & 0.7390      & \textbf{0.7405}              \\\cline{2-14}
                           & Spatter    & 0.5898            & 0.6018      & \textbf{0.6190}              & 0.7411            & 0.7536      & \textbf{0.7797}              & 0.4353            & 0.4756      & \textbf{0.4921}              & 0.7195            & 0.7395      & \textbf{0.7598}              \\ \hline
\multirow{6}{*}{Digital}   & Contrast   & \textbf{0.6307}            & 0.6139      & 0.6255              & 0.7382            & 0.7271      & \textbf{0.7677}              & 0.4364            & \textbf{0.4624}      & 0.4469              & 0.6966            & 0.7240      & \textbf{0.7314}              \\ \cline{2-14}
                           & Elastic    & 0.5959            & 0.6059      & \textbf{0.6226}              & 0.7303            & 0.7465      & \textbf{0.7628}              & 0.4404            & 0.4672      & \textbf{0.5014}              & 0.6906            & 0.7230      & \textbf{0.7355}              \\ \cline{2-14}
                           & Gamma      & 0.6207            & 0.6342      & \textbf{0.6484}              & 0.7782            & 0.7798      & \textbf{0.7923}              & 0.4453            & 0.4683      & \textbf{0.4927}              & 0.7531            & 0.7599      & \textbf{0.7721}              \\ \cline{2-14}
                           & Jpeg       & 0.5935            & 0.6078      & \textbf{0.6195}              & 0.7513            & 0.7565      & \textbf{0.7798}              & 0.4343            & \textbf{0.4956}      & 0.4867              & 0.7530            & 0.7527      & \textbf{0.7734}              \\ \cline{2-14}
                           & Pixelate   & 0.6219            & 0.6311      & \textbf{0.6556}              & 0.7697            & 0.7747      & \textbf{0.7917}              & 0.4538            & 0.4699      & \textbf{0.4802}              & 0.7610            & 0.7600      & \textbf{0.7776}              \\ \cline{2-14}
                           & Saturate   & 0.6095            & 0.6327      & \textbf{0.6437}              & 0.7682            & 0.7684      & \textbf{0.7821}              & 0.4516            & \textbf{0.4757}      & 0.4621              & 0.7552            & 0.7591      & \textbf{0.7699}              \\ \hline
\multicolumn{2}{c|}{Mean}                & 0.6063            & 0.6203      & \textbf{0.6367}     & 0.7568            & 0.7644      & \textbf{0.7814}     & 0.4421            & 0.4715      & \textbf{0.4764}     & 0.7381            & 0.7494      & \textbf{0.7629} \\
\bottomrule[1pt] 
\end{tabular}}
\end{table*}

\subsection{Discussion and Limitations}
Overall, our approach has demonstrated substantial improvements in various comparative experiments. It effectively enhances the performance of question-answering and answer localization tasks and exhibits exceptional adaptability to domain shift and image corruption challenges. This success can be attributed to our proposed 
C$^2$G-ViL module, which aligns multimodal representations and globally calibrates contextual features, thereby enhancing the informative interactions between modalities during feature fusion. Additionally, our exploration of optimal fusion weights and contrastive training strategies enables our models to recognize and capture subtle discrepancies or patterns.

However, we have also observed poor F-Score results for our methods compared to other sophisticated approaches, particularly in certain instances of the VQLA task. This indicates that our work may underperform in specific scenarios. Therefore, as a potential avenue for future research, it would be worthwhile to explore more optimal information fusion strategies to improve the precision and recall of our model while balancing all the relevant metrics.
Besides, considering the increasing demand for specialized query understanding in surgery and surgical procedures, it is necessary to acknowledge the limitation of our current method, which may not fully meet the accuracy requirements of the VQLA task, especially when facing domain shifts. In addition to developing domain adaptation and continual learning strategies, one potential direction for future work could involve exploring gradient perturbation techniques for multimodal fusion features. These advancements would help better handle domain shifts and image corruption challenges while simultaneously enhancing prediction accuracy.

Furthermore, current Multimodal Large Language Models (MLLMs) have become crucial for general multimodal understanding tasks due to their large parameter counts and extensive training data, which enable them to display advanced capabilities~\cite{achiam2023gpt,bai2023qwen}. These models can understand and produce complex responses that mimic human-like interactions. However, MLLMs often respond slowly and require high-performance GPUs for training and deployment. For domain-specific issues, MLLMs need targeted fine-tuning based on specific data, and they may also encounter problems with hallucinations and the repetitive output phenomenon~\cite{gunjal2024detecting,shen2024pmg}. In contrast, Surgical VQLA++ is specifically designed to provide accurate and context-aware responses to queries about particular surgical images, which is vital in medical settings where the correctness of information can greatly influence diagnosis and treatment decisions. In surgical environments, decisions must frequently be made swiftly based on the latest visual information. Surgical-VQLA++ supports real-time decision-making in these critical situations and uses adversarial contrastive learning to improve its robustness against image corruptions, making it suitable for medical applications. Moreover, while the surgical report or caption generation task can only provide general textual descriptive output, our VQLA model can deliver specific answers to given questions and also provide corresponding localization results, enhancing both the precision and utility of visual information processing in surgical contexts.

\section{Conclusion}

In this study, we introduced Surgical-VQLA++, an advanced model that enhances the robustness and VQLA performance in the context of robotic surgery. Our framework integrates C$^2$G-ViL embedding with adversarial contrastive learning, focusing on precise multimodal information alignment and robust feature extraction. The application of these methodologies ensures that our model not only accurately identifies and localizes important surgical details but also remains resilient against real-world image corruptions. Our extensive evaluations across expanded versions of the EndoVis-18-VQLA and EndoVis-17-VQLA datasets have demonstrated significant improvements in both accuracy and robustness compared to prior models. These achievements underscore the potential of Surgical-VQLA++ to assist in surgical education and clinical decision-making by providing more insightful analyses of surgical scenes.

Moving forward, our potential future works aim to further enhance the capabilities of VQLA in surgical settings. We plan to integrate VQLA systems with mixed reality or simulations to improve surgical education, quickly addressing the queries of students and junior surgeons and reducing the reliance on limited teaching resources. Additionally, deploying VQLA systems on surgical robots or navigation tools will offer real-time visual assistance, thereby improving the efficiency and safety of surgical operations. To tackle the challenges in VQLA data collection, such as domain shift and temporal discontinuity, we also aim to incorporate auxiliary information from MLLMs and apply continual learning techniques. This approach will help in fine-tuning well-trained VQLA models across diverse datasets, ensuring high performance on both existing and new data and fostering the development of a unified VQLA framework. These initiatives are expected to enhance prediction accuracy and expand the applications of VQLA in surgical environments.

\section*{Acknowledgement}
This work was supported by Hong Kong Research Grants Council (RGC) Collaborative Research Fund (CRF C4026-21GF), General Research Fund (GRF 14203323, GRF 14216022, and GRF 14211420),  NSFC/RGC Joint Research Scheme N\_CUHK420/22; Shenzhen-Hong Kong-Macau Technology Research Programme (Type C) STIC Grant 202108233000303.
M. Islam was funded by EPSRC Grant [EP/W00805X/1].


\bibliographystyle{model1-num-names}

\bibliography{cas-refs}

\balance

\end{document}